%% file: main.tex
\documentclass[sigconf, nonacm]{acmart}

\newcommand\vldbdoi{XX.XX/XXX.XX}
\newcommand\vldbpages{XXX-XXX}
\newcommand\vldbvolume{20}
\newcommand\vldbissue{1}
\newcommand\vldbyear{2027}
\newcommand\vldbauthors{\authors}
\newcommand\vldbtitle{\shorttitle} 
\newcommand\vldbavailabilityurl{https://github.com/rueckstiess/origami-jsynth}
\newcommand\vldbpagestyle{plain} 

\input{math_commands}

\usepackage{booktabs, multirow}
\usepackage{subcaption}
\usepackage{listings}

\begin{document}
\title{Autoregressive Synthesis of Sparse and Semi-Structured Mixed-Type Data}

\author{Thomas R\"uckstie\ss}
\affiliation{%
  \institution{Independent Researcher}
  \city{Melbourne}
  \state{Australia}
  \postcode{3141}
}
\email{research@tomr.au}

\author{Robin Vujanic}
\affiliation{%
  \institution{MongoDB}
  \city{Sydney}
  \country{Australia}
}
\email{robin.vujanic@mongodb.com}

\begin{abstract}
Synthetic data generation is an important capability for privacy-preserving data sharing, system benchmarking and test data provisioning. For mixed-type data, existing synthesizers largely target dense, fixed-schema tables, but many modern data systems store and exchange sparse, semi-structured JSON with nested objects, variable-length arrays and optional keys. Applying tabular synthesizers to such data requires flattening records into wide, sparse tables, turning nested structure and arrays into column-layout artifacts. We present \origami, an autoregressive transformer architecture for modeling and synthesizing semi-struc\-tured records without flattening. \origami serializes JSON records into key, value, and structural tokens, and encodes token positions by their path in the document tree. Grammar and schema constraints enforce syntactically valid JSON and dataset-consistent structure.
We evaluate \origami against VAE, GAN, diffusion, and autoregressive baselines that operate on flattened representations across six datasets ranging from dense tabular benchmarks to large-scale semi-structured collections. Across fidelity, detection, and utility metrics, \origami achieves the best score in 17 of 18 benchmark comparisons, while maintaining high privacy scores above 96\% across all settings. These results establish native record modeling as a strong alternative to tabular synthesis pipelines, preserving structure while achieving state-of-the-art benchmark performance.
\end{abstract}

\maketitle

\pagestyle{\vldbpagestyle}
\begingroup\small\noindent\raggedright\textbf{PVLDB Reference Format:}\\
\vldbauthors. \vldbtitle. PVLDB, \vldbvolume(\vldbissue): \vldbpages, \vldbyear.\\
\href{https://doi.org/\vldbdoi}{doi:\vldbdoi}
\endgroup
\begingroup
\renewcommand\thefootnote{}\footnote{\noindent
This work is licensed under the Creative Commons BY-NC-ND 4.0 International License. Visit \url{https://creativecommons.org/licenses/by-nc-nd/4.0/} to view a copy of this license. For any use beyond those covered by this license, obtain permission by emailing \href{mailto:info@vldb.org}{info@vldb.org}. Copyright is held by the owner/author(s). Publication rights licensed to the VLDB Endowment. \\
\raggedright Proceedings of the VLDB Endowment, Vol. \vldbvolume, No. \vldbissue\ %
ISSN 2150-8097. \\
\href{https://doi.org/\vldbdoi}{doi:\vldbdoi} \\
}\addtocounter{footnote}{-1}\endgroup

\ifdefempty{\vldbavailabilityurl}{}{
\vspace{.3cm}
\begingroup\small\noindent\raggedright\textbf{PVLDB Artifact Availability:}\\
The source code, data, and/or other artifacts have been made available at \url{\vldbavailabilityurl}.
\endgroup
}

\input{01_introduction}
\input{02_related_work}
\input{03_architecture}
\input{04_experiments}

\input{05_discussion}
\input{06_conclusion}


\bibliographystyle{ACM-Reference-Format}
\bibliography{references}

\end{document}

%% file: math_commands.tex
\usepackage{amsmath,amsfonts,bm}
\usepackage{xspace}

\newcommand{\origami}{\textsc{origami}\xspace}

\def\eqref#1{equation~\ref{#1}}

\def\1{\bm{1}}

\DeclareMathAlphabet{\mathsfit}{\encodingdefault}{\sfdefault}{m}{sl}
\SetMathAlphabet{\mathsfit}{bold}{\encodingdefault}{\sfdefault}{bx}{n}



%% file: 01_introduction.tex
\section{Introduction}

Synthetic data generation has become an important capability in modern data
management. Organizations require realistic data for privacy-preserving
data sharing, software testing, provisioning of development and
QA environments, training of Machine Learning models and benchmarking database workloads
at scale~\cite{jordonSyntheticDataWhat2022,schmidtSQLStormTakingDatabase2025}. 

A rich body of work now addresses mixed-type data synthesis \cite{shiComprehensiveSurveySynthetic2025}, 
spanning generative adversarial networks (GAN) and
variational autoencoders (VAE) ~\cite{xuModelingTabularData2019, liu2023goggle},
diffusion models~\cite{kotelnikovTabDDPMModellingTabular2024, zhangMixedTypeTabularData2024, shiTabDiffMixedtypeDiffusion2025},
and autoregressive approaches~\cite{solatorioREaLTabFormerGeneratingRealistic2023, borisovLanguageModelsAre2023, crompTabbyLanguageModel2026, tiwaldTabularARGNFlexibleEfficient2025a}. 
Despite this architectural diversity, one assumption has remained
largely unchallenged: all methods operate on fixed-schema tables
with dense, homogeneously typed columns, treating
sparsity  as a data issue that needs to be fixed during 
preprocessing. Approaches to tackle this sparsity
range from mean imputation of missing numerics, sentinel values
for categorical missingness, to dropping sparse rows entirely---in 
each case destroying structural properties of the data that
we argue ought to be modeled.

This \emph{dense table} assumption is increasingly at odds with a large and growing
share of modern data workloads. While transactional systems continue
to rely on relational schemas, application-layer data is
overwhelmingly exchanged and often stored in semi-struc\-tured formats.
Document databases, REST APIs, data lakes, and event streams operate
on JSON records rather than flat tables~\cite{carey2025principled,stonebraker2024goes,bourhisJSONDataModel2017a}. A single business listing in
a review platform, for instance, may contain nested objects for
operating hours and location, variable-length arrays for categories
and reviews, optional keys that differ across records, and even type
polymorphism where the same key path holds an integer in one record
and a string in another. 

To apply existing data synthesis methods to semi-structured data, 
the only recourse is to first flatten nested structures into tables. 
But this transformation is problematic and scales poorly. 
Variable-length arrays produce sparse trailing columns,
type-polymorphic keys require splitting into per-type sub-columns,
and optional keys carry semantic meaning that is structural, not stochastic. 
In our experiments, flattening real-world JSON datasets
produces tables with many hundreds of columns and sparsity exceeding
93\%. These tables are qualitatively different from the dense 10--20 column 
benchmark datasets from the UCI repository~\cite{kellyUCIMLRepo} 
frequently found in the synthetic data generation literature,
e.g. Adult (48K rows, 15 columns), Shoppers (12K rows, 18 columns), and
Magic (19K rows, 11 columns). 

A secondary tension concerns the treatment of mixed types. Methods that
operate in continuous latent space (GANs, VAEs, diffusion models) handle
numerics natively but must encode categorical columns, often through
one-hot vectors that scale poorly with cardinality. This failure mode 
gets amplified in flattened semi-structured data, as array expansion 
can inflate the categorical column count into the hundreds or thousands. 
On the other hand, autoregressive and LLM-based methods handle categoricals natively
as tokens but must discretize high-cardinality numeric values to avoid
vocabulary explosion, which sacrifices precision and ordinal structure.

To address these issues, we present \origami (Object RepresentatIon 
via Generative Autoregressive ModelIng), an autoregressive transformer 
with a dual-head discrete/continuous architecture that operates 
directly on tokenized JSON records, avoiding flattening entirely.
Our tokenization scheme serializes nested objects, arrays,
and primitive values into sequences of key, value, and structural
tokens, preserving hierarchy and sparsity as first-class properties of
the data. Key-Value Position Encoding (KVPE) replaces the usual sequential
position indices with structural path encodings derived from each
token's location in the record tree, making the model invariant to the
order of sibling keys. This invariance enables key-order shuffling as a
data augmentation strategy that prevents memorization by presenting
each training record in a different key order at every epoch.
The dual-head architecture predicts discrete tokens (keys, categorical
values, and structural delimiters) through next-token prediction, while
continuous numeric values are modeled as parameterized Mixture of Gaussians, 
handling both data types in their native representation without discretization or
one-hot encoding. Grammar and schema constraints, enforced via a
pushdown automaton and a compiled mask table, guarantee that every
generated record is syntactically valid JSON conforming to the derived
data schema.

Although \origami is designed for semi-structured data modeling, it also
achieves excellent results on standard tabular benchmarks, which are
representable as flat JSON key-value records. Across six datasets,
including sparse semi-structured collections with up to 93\% key-path sparsity
after flattening and over one million records, \origami achieves the
best score in 17 of 18 fidelity, detection, and utility comparisons
against VAE, GAN, diffusion, and autoregressive baselines, while
retaining privacy scores above 96\%. 

Our contributions in this work are as follows:
\begin{enumerate}
  \item We introduce \origami, a purpose-built autoregressive architecture for
    synthesizing semi-structured records directly in their JSON
    representation, natively handling hierarchical nesting,
    variable-length arrays, sparsity, and type polymorphism without
    flattening or imputation.

  \item We propose Key-Value Position Encoding (KVPE), which represents
    tokens by their structural path in the record tree, enabling
    order-invariant modeling of shuffled key-value data and reducing
    memorization through key-order augmentation.

  \item We develop a flattening and type-separation methodology, along
    with extensions to standard tabular metrics, for evaluating
    semi-structured synthesizers against tabular baselines while
    preserving type fidelity and structural missingness patterns.

    \item We evaluate \origami on six datasets ranging from standard
      tabular benchmarks to large-scale JSON-native collections with over
      one million records, showing consistently strong fidelity, detection,
      utility, and privacy results across both dense tabular and sparse
      semi-structured settings.
\end{enumerate}
\noindent All datasets and code to reproduce our experiments is available\footnote{\url{https://github.com/rueckstiess/origami-jsynth}}. \origami is also released as a Python package (\texttt{origami-ml}) under permissive Apache 2.0 license with SDK and CLI interfaces.

%% file: 02_related_work.tex
\section{Related Work}

\paragraph{GAN and VAE-based Synthesis}
Generative Adversarial Networks \citep{goodfellowGenerativeAdversarialNetworks2014}
train a generator to produce synthetic samples that a co-trained discriminator
cannot distinguish from real data.
Variational Autoencoders \cite{kingmaAutoEncodingVariationalBayes2014} instead
learn a regularized latent space from which new samples can be decoded, optimizing
a lower bound on the data likelihood.
Both paradigms have been adapted for tabular synthesis, most notably as
CTGAN and TVAE \cite{xuModelingTabularData2019}. Many variants have since been proposed~\cite{shiComprehensiveSurveySynthetic2025}, e.g.
GOGGLE~\cite{liu2023goggle}, which extends the VAE framework by replacing the MLP decoder with a Message Passing Neural Network over a jointly learned column
dependency graph.

\paragraph{Diffusion Synthesis}
Diffusion models generate data by learning to reverse a noise process that gradually
corrupts real samples into Gaussian noise. TabDDPM \cite{kotelnikovTabDDPMModellingTabular2024} 
first applied diffusion to mixed-type tabular data via separate Gaussian and multinomial
processes for numerical and categorical features. TabSyn \cite{zhangMixedTypeTabularData2024}
improved sampling speed and fidelity by performing diffusion in a VAE latent space, and 
TabDiff \cite{shiTabDiffMixedtypeDiffusion2025} further introduced feature-wise learnable
noise schedules to adaptively allocate model capacity across heterogeneous columns, achieving
state-of-the-art results on many standard benchmarks.

\paragraph{Autoregressive Synthesis}
Autoregressive models decompose the joint distribution into a product of conditionals, 
$p(x_1, \ldots, x_n) = \prod_{i=1}^{n} p(x_i \mid x_1, \ldots, x_{i-1})$, generating
one variable at a time. GReaT \cite{borisovLanguageModelsAre2023} and Tabby
\cite{crompTabbyLanguageModel2026} fine-tune a pretrained GPT-2
\cite{radfordImprovingLanguageUnderstanding2018} backbone on rows serialized as natural
language, with Tabby adding a Mixture-of-Experts output head per column. REaLTabFormer
\cite{solatorioREaLTabFormerGeneratingRealistic2023} instead trains a GPT-2 architecture
from scratch on a dataset-specific vocabulary with digit-level numerical encoding and
constrained decoding. \origami's approach is similar in that is also produces a
dataset-specific vocabulary of keys and values, but additionally includes structural
tokens to model hierarchy and nesting. TabularARGN \citep{tiwaldTabularARGNFlexibleEfficient2025a}
departs from transformers, adopting a lightweight NADE-inspired \citep{uriaDeepTractableDensity2014}
architecture of per-column MLP regressors over discretized sub-columns. Several autoregressive
methods randomly permute the factorization order during training as proposed 
by~\cite{yangXLNetGeneralizedAutoregressive2020, alcornDEformerOrderAgnosticDistribution2021},
effectively learning an ensemble over all orderings to regularize against spurious
correlations and enable arbitrary conditional generation at inference, which \origami employs as well.

While not specifically designed for data synthesis, \cite{golkarXValContinuousNumber2023} adds a second
continuous head to the standard LLM architecture, making it a hybrid discrete/continuous model.
We adopt this approach in \origami, however instead training with MSE loss for single
scalar regression, we model continuous values as a Mixture of Gaussians and train with a
negative log likelihood loss to better adapt to uncertainty and multi-modal distributions. 
For constrained decoding, we extend ideas from \cite{willardEfficientGuidedGeneration2023,kooAutomatabasedConstraintsLanguage2024},
applying grammar and schema constraints during both training and inference to accelerate convergence
and enforce valid token sequences.

\paragraph{Synthesis beyond Single Tables}
The database community has explored synthetic data generation for
benchmarking and privacy-preserving data sharing.
SAM~\cite{yangSAMDatabaseGeneration2022} uses supervised autoregressive
models to generate databases satisfying cardinality constraints from
query workloads, and PrivBench~\cite{gePrivBenchPrivacyenhancedDatabase2025}
synthesizes benchmark databases that preserve both data distributions
and query runtime characteristics under differential privacy (DP).
For multi-table settings,
PrivLava~\cite{caiPrivLavaSynthesizingRelational2023} introduces
graphical models with latent variables to capture inter-table
correlations under DP, and
REaLTabFormer~\cite{solatorioREaLTabFormerGeneratingRealistic2023}
extends its single-table transformer with a Seq2Seq model for
parent--child table relationships.
All of these methods assume relational schemas with fixed-width
rows; none operate on semi-structured records with variable
nesting or optional keys.

\paragraph{Sparsity and Missing Data}
Most tabular synthesizers assume fully observed inputs and handle
missingness through imputation or row dropping.
\cite{mohapatraDifferentiallyPrivateData2024} formalize DP synthetic
data generation when inputs contain missing values, proposing
adaptive strategies that improve utility by modeling missingness
patterns rather than imputing them away.
In scientific computing, Sparse Data
Diffusion~\cite{ostheimerSparseDataDiffusion2025} explicitly encodes
exact zeros through auxiliary ``sparsity bits'' rather than treating
them as continuous values, recognizing that sparsity carries semantic
information.
Both works identify the need to model sparsity as a first-class
property of the data, but neither extends to the hierarchical,
variable-schema setting that characterizes semi-structured data.

%% file: 03_architecture.tex
\section{Architecture}

\origami falls into the category of autoregressive transformer-based methods. 
Unlike previous work, which assumes flat tabular inputs, it operates on tokenized
representations of JSON records. We describe each component in turn:
preprocessing, tokenization, the model architecture including Key-Value
Position Encoding (KVPE), output heads, grammar and schema constraints
and post-processing of numeric data.

Throughout this work, we use standard tabular terminology, \emph{rows}
and \emph{columns}, when referring to flat table representations. For
semi-structured data, we use \emph{records} (analogous to rows) and
\emph{keys} (analogous to columns), to refer to the entries and named
attributes of a JSON record. A \emph{key path} is the dot-separated
concatenation of nested keys through the record tree, with integer indices
for array positions: for example, \texttt{user.addresses.0.city} refers
to the \texttt{city} key of the first element in the \texttt{addresses}
array under the \texttt{user} key.

\subsection{Preprocessing}
\label{sec:preprocessing}

Numeric keys in semi-structured data span a wide range of
cardinalities. Low-cardinality keys (e.g., a \texttt{rating} key
with integers $1$--$5$) are treated as categorical tokens. For
high-cardinality numeric keys (more than $\tau$ distinct values, where
$\tau$ is a configurable threshold), we apply per-key standardization.
Let $x$ be a value of key $k$; we compute
\begin{equation}
  \tilde{x} = \frac{x - \mu_k}{\sigma_k},
\end{equation}
where $\mu_k$ and $\sigma_k$ are the mean and standard deviation of
numeric values under key $k$ estimated from training data. Scaled values are passed to the
model through a dedicated continuous channel described in
Section~\ref{sec:continuous-head} and inverse-transformed during post-processing (Section~\ref{sec:postprocessing}).

\subsection{Tokenization}
\label{sec:tokenization}

Our tokenization scheme differs from typical tokenizers used in the
language modeling literature, such as Byte-pair Encoding
\cite{gageNewAlgorithmData1994} and WordPiece \cite{wuGooglesNeuralMachine2016},
in that we tokenize JSON instances into key and value tokens and maintain
the structure of arrays and nested objects with special grammatical tokens.

For a dataset $\mathcal{D}$, each JSON record $d \in \mathcal{D}$ is serialized into a token sequence
$\mathbf{x} = (x_1, \ldots, x_T)$ by a depth-first traversal. The
vocabulary $\mathcal{V}$ consists of three disjoint token classes,
$\mathcal{V} = \mathcal{V}_s \cup \mathcal{V}_k \cup \mathcal{V}_v$
as follows:
\begin{itemize}
  \item \textbf{Structural tokens} $\mathcal{V}_s = \{\textsc{start},
    \textsc{end}, \textsc{obj\_start}, \textsc{obj\_end},\\
    \textsc{arr\_start}, \textsc{arr\_end}, \textsc{pad},
    \textsc{num}\}$, which delimit record boundaries, objects, arrays
    and a sentinel NUM placeholder when continuous modeling is enabled.
  \item \textbf{Key tokens} $\mathcal{V}_k$, one per distinct key
    observed in training.
  \item \textbf{Value tokens} $\mathcal{V}_v$, one per
    distinct categorical value (strings, booleans, \texttt{null}, and
    low-cardinality numbers).
\end{itemize}

Nested objects and arrays are handled recursively: an object value emits
$\textsc{obj\_start} \ldots \textsc{obj\_end}$, and an array emits
$\textsc{arr\_start} \ldots$ $\textsc{arr\_end}$ with its elements in
order.

High-cardinality numeric values that were standardized in preprocessing (Section~\ref{sec:preprocessing}) 
emit a special \textsc{num} token in the discrete sequence, with the
scaled value $\tilde{x}$ stored in a parallel continuous channel to train the continuous head (Section~\ref{sec:continuous-head}).

Alongside each token $x_t$, the tokenizer records its key path
$\mathbf{p}_t = (e_1, \ldots, e_{D_t})$ through the JSON hierarchy,
where each element $e_i$ is either a key name or an array index, and
$D_t$ is the nesting depth. For instance, in
\texttt{\{"user":\{"name":"Alice"\}\}}, the token for value
\texttt{Alice} has path $(\texttt{Key(user)},\;
\texttt{Key(name)})$.

\begin{figure}
  \centering
  \includegraphics[width=1.05\linewidth]{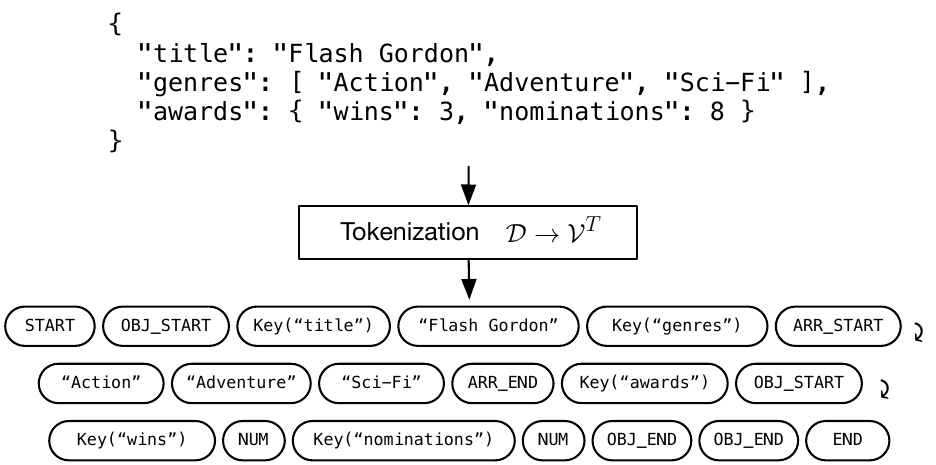}
  \caption{Tokenization of an example \textit{movies} record.}
  \label{fig:preprocessing}
\end{figure}

Figure~\ref{fig:preprocessing} illustrates the tokenization of an
example JSON record.

\subsection{Input Representation}

The input to the transformer at position $t$ is the sum of a token
embedding and a position embedding:
\begin{equation}
  \mathbf{h}_t^{(0)} = \mathbf{e}(x_t) + \text{KVPE}(\mathbf{p}_t),
\end{equation}
where $\mathbf{e}: \mathcal{V} \to \mathbb{R}^{d}$ is a learned token
embedding.

\paragraph{Key-Value Position Encoding (KVPE)}
Standard transformers typically use sequential, sinusodal \cite{vaswaniAttentionAllYou2017} or
rotary \cite{suRoFormerEnhancedTransformer2023a} position indices.
Since JSON key-value pairs have no inherent order, these positions would
impose a spurious ordering on sibling keys. Instead, KVPE encodes
each token's \emph{structural position} representing its path through the
record tree.

Each path element is embedded independently: key elements reuse the
token embedding matrix $\mathbf{e}$ (tying key representations across
the position and content channels), while array index elements use a
separate embedding table
$\mathbf{e}_{\mathrm{idx}}: \{0, \ldots, I_{\max}\} \to
\mathbb{R}^{d}$, where $I_{\max}$ is a fixed, configurable capacity
for array position embeddings. The sequence of element embeddings
$(\mathbf{e}(e_1), \ldots, \mathbf{e}(e_{D_t}))$ is aggregated into a
single position vector via sum pooling: $\text{KVPE}(\mathbf{p}_t) = \sum_{i=1}^{D_t} \mathbf{e}(e_i)$.


\paragraph{Numeric embedding}
For positions where $x_t = \textsc{num}$,
we follow the approach in \cite{golkarXValContinuousNumber2023}
and replace the token embedding $\mathbf{e}(x_t)$ with a multiplicative embedding
$\tilde{x}_t \cdot \mathbf{v}_{\mathrm{num}}$, where
$\mathbf{v}_{\mathrm{num}} \in \mathbb{R}^d$ is a learned direction
vector and $\tilde{x}_t$ is the standardized value. This injects
continuous numeric information directly into the
representation, encoding both the sign and magnitude of the
standardized value as the direction and norm of the embedding vector.

\subsection{Transformer Backbone}

The backbone is a stack of $L$ pre-norm decoder-only transformer
layers with causal (autoregressive) attention:
\begin{align}
  \mathbf{z}_t^{(\ell)} &= \mathbf{h}_t^{(\ell)} +
    \text{MHA}\bigl(\text{LN}(\mathbf{h}_t^{(\ell)})\bigr), \\
  \mathbf{h}_t^{(\ell+1)} &= \mathbf{z}_t^{(\ell)} +
    \text{FFN}\bigl(\text{LN}(\mathbf{z}_t^{(\ell)})\bigr),
\end{align}
where $\text{MHA}$ is multi-head attention with $H$ heads and head
dimension $d/H$, $\text{FFN}$ is a two-layer feed-forward network with
GELU~\cite{hendrycks2016gaussian} activation and hidden dimension $d_{\mathrm{ff}}$, and
$\text{LN}$ denotes layer normalization. A final layer norm is applied
after the last layer. Causal masking ensures each position attends only
to itself and earlier positions.

\paragraph{Left-padding.}
Since JSON records vary in length, batches are \emph{left-padded} so
that all sequences end at the same position. This allows
$\mathbf{h}_{T}^{(L)}$ to always be the representation of the last
real token, simplifying batched next-token prediction.

\subsection{Output Heads}

\begin{figure}[tb]
    \centering
    \includegraphics[width=\columnwidth]{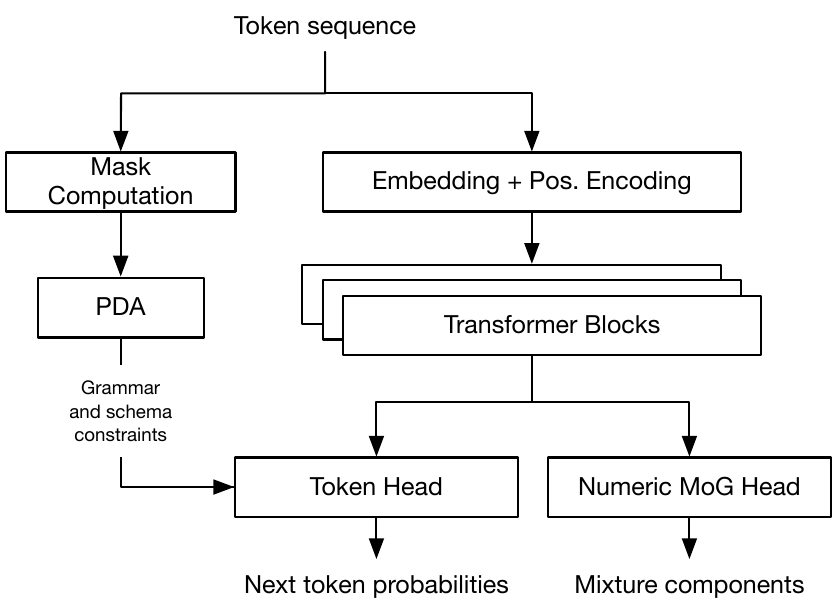}
    \caption{\origami dual-head model architecture with grammar and schema constraints imposed on the discrete head.}
    \label{fig:architecture}
\end{figure}

The model has two output heads, inspired by~\cite{golkarXValContinuousNumber2023}: a discrete head for structural, key,
and categorical value tokens, and an optional continuous head for
numeric values, as shown in Figure~\ref{fig:architecture}.

\paragraph{Discrete head.}
A linear projection maps the final hidden state to vocabulary logits:
\begin{equation}
  \ell_t = \mathbf{W}_d \, \mathbf{h}_t^{(L)} + \mathbf{b}_d
  \in \mathbb{R}^{|\mathcal{V}|},
\end{equation}
trained with cross-entropy loss over all non-padding positions.

\paragraph{Continuous head (Mixture of Gaussians).}
\label{sec:continuous-head}
For high-cardinality numeric keys, the discrete head emits a
\textsc{num} token while a parallel continuous head models the
numeric value. The continuous head projects the hidden state to
parameters of a Mixture of Gaussians (MoG) with $K$ components:
\begin{equation}
  (\boldsymbol{\pi}_t,\; \boldsymbol{\mu}_t,\;
  \log\boldsymbol{\sigma}^2_t)
  = \text{split}\bigl(\mathbf{W}_c \, \mathbf{h}_t^{(L)} +
  \mathbf{b}_c\bigr)
  \in \mathbb{R}^{3K},
\end{equation}
where $\boldsymbol{\pi}_t = \text{softmax}(\cdot)$ are mixture weights.
The numeric value $\tilde{x}_{t+1}$ is modeled as:
\begin{equation}
  p(\tilde{x}_{t+1} \mid \mathbf{h}_t^{(L)}) = \sum_{j=1}^{K}
  \pi_{t,j}\; \mathcal{N}\bigl(\tilde{x}_{t+1};\; \mu_{t,j},\;
  \sigma^2_{t,j}\bigr).
\end{equation}
The continuous loss is the negative log-likelihood over positions where
the target is a \textsc{num} token. The total training loss is:
\begin{equation}
  \mathcal{L} = \mathcal{L}_{\mathrm{CE}} +
  \lambda \, \mathcal{L}_{\mathrm{NLL}},
\end{equation}
where $\lambda$ is set proportionally to the fraction of \textsc{num}
tokens to total tokens in each batch.

\subsection{Key-Order Shuffling}

Since JSON object keys are unordered by specification, we randomly
permute the key order at each nesting level every time a training
example is accessed. Combined with KVPE (which encodes structural
position rather than sequential position), this augmentation forces the
model to attend to key semantics rather than memorizing a canonical
ordering. We demonstrate through ablation (see Section~\ref{sec:discussion} and Figures~\ref{fig:kvpe-ablation} and \ref{fig:shuffle-ablation}) that this approach helps distinguish robust causal relationships
from spurious correlations, as truly meaningful dependencies will
persist across different ordering permutations, while coincidental 
correlations tend to average out during training. This mechanism also 
serves as an implicit regularization strategy similar to Dropout 
\cite{srivastavaDropoutSimpleWay2014}, as the model must learn to 
predict from varying subsets of conditioning variables. However, 
unlike Dropout where a portion of the gradient information is 
discarded, the order-agnostic approach maintains full use of the 
training signal.

\subsection{Grammar Constraints}
\label{sec:grammar}

To guarantee syntactically valid JSON output, we employ a pushdown
automaton (PDA) that tracks the grammatical state during both training
and generation. The PDA maintains a stack encoding the current nesting
context (object vs.\ array at each depth level) along with flags for
parser state (e.g., whether a key is awaiting its value).

At each position $t$, the PDA computes a boolean mask
$\mathbf{m}_t \in \{0,1\}^{|\mathcal{V}|}$ indicating which tokens are
grammatically valid as the next token. Invalid token logits are set to
$-\infty$ before the softmax:
\begin{equation}
  \hat{\ell}_{t,v} =
  \begin{cases}
    \ell_{t,v} & \text{if } m_{t,v} = 1, \\
    -\infty    & \text{otherwise.}
  \end{cases}
\end{equation}

During training, grammar masks are computed in parallel across all
positions. Since the PDA state updates involve many small sequential
operations that incur synchronization overhead on GPU, mask computation
is offloaded to CPU workers in the data loading pipeline: while the GPU
executes forward and backward passes on the current batch, DataLoader
workers prepare grammar masks for subsequent batches in
parallel, hiding the constraint computation cost behind GPU-bound
training. During autoregressive generation, the PDA state is updated
incrementally in $O(1)$ per step.

\subsection{Schema Constraints}
\label{sec:schema}

The grammar PDA (Section~\ref{sec:grammar}) enforces syntactic
validity but says nothing about the \emph{semantic} structure of the
data: which keys may appear, what types each key admits, or which
values are legal. To enforce these constraints, we derive a JSON
Schema \cite{pezoa2016foundations} (draft 2020-12 subset) from the training data and compile it into a schema mask that
is intersected with the grammar mask at every decoding position.

\paragraph{Schema derivation.}
Given a training corpus $\mathcal{D}$, we automatically derive a JSON
Schema by analyzing the data structure:
\begin{itemize}
  \item \textbf{Types}: Each key's observed Python types are mapped
    to JSON Schema types (\texttt{string}, \texttt{integer},
    \texttt{number}, \texttt{boolean}, \texttt{null}, \texttt{object},
    \texttt{array}).
  \item \textbf{Enumerations}: Keys with at most $\tau$ distinct
    primitive values receive an \texttt{enum} constraint listing
    all observed values.
  \item \textbf{Key restrictions}: Object schemas set \\
    \texttt{additionalProperties:false}, restricting keys to those
    observed in training. Keys present in every object are marked
    \texttt{required}.
  \item \textbf{Array bounds}: Observed array lengths yield
    \texttt{minItems} and  \texttt{maxItems} constraints. Arrays where all
    observed instances contain unique elements are marked
    \texttt{uniqueItems}.
  \item \textbf{Numeric bounds}: Observed \texttt{minimum} and
    \texttt{maximum} values are recorded per numeric key.
\end{itemize}
When the continuous numeric mode is active, the schema is transformed
to reflect the preprocessed representation: enum values for scaled
keys are removed (since numeric values are now continuous) and
bounds are mapped to the standardized scale.

\paragraph{Compiled mask table.}
The schema is compiled into a mask table
$\mathbf{M} \in \{0,1\}^{(P+1) \times |\mathcal{V}|}$, where $P$ is
the number of unique key paths. Row~0 is all-ones (the default for
positions outside the schema, such as record delimiters);
each subsequent row $i$ is a boolean mask reflecting the type, enum,
and key restrictions for key path~$i$. Each token position $t$ in
the JSON tree maps to a key path via its KVPE path
$\mathbf{p}_{t+1}$ (with array indices replaced by a wildcard
$\ast$). The schema mask for the full sequence is produced by mapping each
position's path to its schema table row and performing a single
gather operation. The effective constraint mask is the intersection
of grammar and schema masks:
$\hat{\mathbf{m}}_t = \mathbf{m}_t \wedge \mathbf{s}_t$.

This design separates \emph{path-dependent} constraints (type, enum,
allowed keys), which are pre-computed and applied via a single tensor
gather at $O(1)$ cost per position, from \emph{count-dependent}
constraints (\texttt{minItems}, \texttt{maxItems}, \texttt{required},
\texttt{uniqueItems}), which require tracking state during generation
and are enforced incrementally at inference time only for performance
reasons.

\subsection{Post-Processing}
\label{sec:postprocessing}

Numeric preprocessing (Section~\ref{sec:tokenization}) introduces
artifacts: standardized values decoded through the continuous head may
not lie on the original scale's natural grid (e.g., integer keys
produce floats, and sampled values may fall outside observed bounds).
We apply a deterministic post-processing pass to each generated value
using the \emph{original-data} schema (before preprocessing
transforms):
\begin{enumerate}
  \item \textbf{Clip to bounds}: Enforce the key's observed
    \texttt{minimum} and \texttt{maximum}.
  \item \textbf{Snap to enum}: If the key has an \texttt{enum}
      constraint, replace the value with the nearest observed value
      for that key by absolute difference.
  \item \textbf{Round to integer}: If the key type is
    \texttt{integer} and no enum applies, round to the nearest integer.
\end{enumerate}
This pipeline is applied recursively to nested objects and arrays.
Post-processing is a lightweight operation that
does not require model inference, and ensures generated data
conforms to the original data's type and domain constraints.

%% file: 04_experiments.tex
\section{Experiments}

\origami is evaluated against six tabular baseline synthesizers on six datasets
spanning dense tabular benchmarks to large-scale semi-structured collections. 
We emphasize that the comparison on semi-structured datasets is not between 
architectures in isolation, but between two paradigms: flatten-then-synthesize
versus native semi-structured generation. The flattening pipeline 
(Section~\ref{sec:tabular-repr})
represents a reasonable approach for applying existing methods to
semi-structured data, as any practitioner facing this problem today would
need to perform a similar transformation.

Section~\ref{sec:tabular-repr} describes how we flatten and type-separate
semi-structured data to enable baseline training and consistent evaluation.
We then introduce the datasets (Section~\ref{sec:datasets}), describe the
experiment protocol (Section~\ref{sec:protocol}) and evaluation metrics (Section~\ref{sec:metrics}), 
and detail per-model training configurations (Section~\ref{sec:baselines}).

\subsection{Tabular Representation of Semi-Structured Data}
\label{sec:tabular-repr}

Existing tabular synthesizers and evaluation metrics require fixed-schema
tables. We apply a two-stage transformation to native JSON data to produce 
a flat table with homogeneously typed columns suitable for both baseline 
training and metric computation. The transformation is illustrated in Figure~\ref{fig:type-separation}.

\begin{figure*}[tb]
\centering

\begin{verbatim}
{"title": "Flash Gordon", "genres": ["Action", "Adventure", "Sci-Fi"], "awards": {"wins": 3, "nominations": 8}}
{"title": "Tron", "genres": ["Action", "Sci-Fi"], "awards": {"wins": "unknown"}}
\end{verbatim}

\medskip
\centerline{$\big\downarrow$\;\; \textsc{Flatten}}
\medskip

\centering

\begin{tabular}{llllll}
\toprule
\texttt{title} & \texttt{genres.0} & \texttt{genres.1} & \texttt{genres.2}
  & \texttt{awards.wins} & \texttt{awards.nominations} \\
\midrule
Flash Gordon & Action & Adventure & Sci-Fi & 3 & 8 \\
Tron & Action & Sci-Fi
  & \textcolor{black!35}{\texttt{NaN}}
  & ``unknown''
  & \textcolor{black!35}{\texttt{NaN}} \\
\bottomrule
\end{tabular}

\medskip
\medskip

\centerline{$\big\downarrow$\;\; \textsc{Separate Types}}
\medskip

\centering
\begin{tabular}{lll cc ccc cc}
\toprule
& & & \multicolumn{2}{c}{\texttt{genres.2}}
    & \multicolumn{3}{c}{\texttt{awards.wins}}
    & \multicolumn{2}{c}{\texttt{awards.nominations}} \\
\cmidrule(lr){4-5} \cmidrule(lr){6-8} \cmidrule(lr){9-10}
\texttt{title} & \texttt{genres.0} & \texttt{genres.1}
  & \texttt{.dtype} & \texttt{.cat}
  & \texttt{.dtype} & \texttt{.num} & \texttt{.cat}
  & \texttt{.dtype} & \texttt{.num} \\
\midrule
Flash Gordon & Action & Adventure
  & \texttt{cat} & Sci-Fi
  & \texttt{num} & 3 & \textcolor{black!35}{NaN}
  & \texttt{num} & 8 \\
Tron & Action & Sci-Fi
  & \texttt{missing} & \textcolor{black!35}{NaN}
  & \texttt{cat} & \textcolor{black!35}{NaN} & unknown
  & \texttt{missing} & \textcolor{black!35}{NaN} \\
\bottomrule
\end{tabular}

\caption{Flattening and type separation of two movie records.
Nested objects and arrays are mapped to dot-separated columns;
variable-length arrays and absent keys produce \texttt{NaN}.
Mixed-type columns (\texttt{awards.wins}: integer vs.\ string) and
partially-present columns are expanded into a type indicator
(\texttt{.dtype}) and per-type value columns.
Homogeneous, fully-present columns
(\texttt{title}, \texttt{genres.0}, \texttt{genres.1})
pass through unchanged.}
\label{fig:type-separation}
\end{figure*}

\paragraph{Flattening.}
Each JSON record is traversed depth-first. Nested keys are concatenated
with dot separators to form column names: key \texttt{``name''} inside
object \texttt{``user''} becomes column \texttt{``user.name''}. Array elements
are indexed numerically: \texttt{``tags.0''}, \texttt{``tags.1''}, etc. Only
leaf values (primitives and nulls) are retained. Since arrays vary in
length across records, shorter instances produce \texttt{NaN} entries in
trailing index columns. The resulting table has one column per unique
leaf key path observed in the dataset. For datasets without nested key paths 
or arrays, flattening produces the original table unchanged.

\paragraph{Type separation.}
After flattening, a single column may contain values of different types
across rows, or be absent from some records entirely (\texttt{NaN}). Such columns are
expanded into typed sub-columns: a categorical indicator column
\texttt{``col.dtype''} records the per-row type (\texttt{num},
\texttt{cat}, \texttt{bool}, \texttt{null}, or \texttt{missing}), and a
separate column per observed type (\texttt{``col.num''}, \texttt{``col.cat''},
\texttt{``col.bool''}) holds the cast value, with \texttt{NaN} in rows of
other types. This preserves the distinction between explicit null values
(key present with value \texttt{null}) and structurally absent keys
(key not present in the record). The explicit modeling of missing values via the \texttt{.dtype} column also allows tabular synthesizers to produce missing values during generation, even if not natively supported. 

For baseline training, type separation is applied selectively: col\-umns
that are already homogeneous and fully present are kept as single
columns to avoid unnecessary expansion. For evaluation, type separation
is applied to all columns unconditionally and includes an additional
\texttt{``col.alen''} column for arrays, tracking their lengths explicitly, so that 
type fidelity and structural patterns can be measured directly. 

\paragraph{Application.}
Baseline synthesizers train on the flattened, type-separated table.
The transformation is then inverted to reconstruct JSON
records. The same flattening and type separation is applied jointly to
real and synthetic records during evaluation, ensuring a consistent
column schema for all metrics. For tabular datasets, flattening is a
no-op and type separation only activates for columns that require it, so
the transformation reduces to the identity for dense, single-type columns.

\subsection{Datasets}
\label{sec:datasets}

We evaluate all baselines on 6 datasets with varying size, categorical
complexity and sparsity: Adult, Diabetes, Electric Vehicles,
Yelp, DDXPlus and Github Issues.

\paragraph{Adult and Diabetes.} These two datasets were chosen as the 
two largest datasets evaluated in TabDiff~\cite{shiTabDiffMixedtypeDiffusion2025}. 
They originally stem from 
the UCI Machine Learning Repository \cite{kellyUCIMLRepo}. We use
Tab\-Diff's exact preprocessing code and splits for reproducibility
and to establish a baseline. We note that TabDiff's preprocessing
of the Diabetes datasets consolidates empty strings (\texttt{""}), 
question marks (\texttt{"?"}) and single spaces (\texttt{" "}) into
a single ``\texttt{nan}'' string and uses an Ordinal Encoder from
scikit-learn \cite{pedregosaScikitlearnMachineLearning2011} to 
process the \texttt{age} column. While \origami technically doesn't 
require this preprocessing, nevertheless we follow this approach for 
the Diabetes dataset. For evaluating ML utility, we predict \texttt{income}
on the Adult dataset and \texttt{readmitted} on Diabetes.

\paragraph{Electric Vehicles.} This tabular dataset contains 210{,}011
electric vehicle registrations from the New York State Department of
Motor Vehicles\footnote{Snapshot obtained on 2026-02-12 from https://data.ny.gov/Transportation/Electric-Vehicle-Registrations/3vp6-cxmr}.
We remove two degenerate columns (VIN is unique and Fuel Type is constant across all records) and convert date
columns to Unix epoch integers. Several numeric keys (Unladen Weight,
Maximum Gross Weight, Passengers) are structurally sparse---absent
for vehicle subtypes where they do not apply---yielding 11\% total
sparsity. The utility target is \texttt{Body Type}.

\paragraph{Yelp.} This JSON-native dataset contains 150{,}346 business
records from the Yelp Open Dataset\footnote{Data obtained from https://www.yelp.com/dataset}.
We remove unique identifier keys and split the comma-separated categories
string into an array. Each record contains nested objects for business
attributes (up to 39 optional keys) and operating hours, plus a
variable-length category array. The \texttt{attributes} object is the
primary source of structural sparsity, varying widely across business types.
After flattening, the dataset expands to 142 columns with 78\% sparsity.
The utility classification target is \texttt{is\_open}.

\paragraph{DDXPlus.}
DDXPlus~\cite{fansitchangoDdxplusNewDataset2022} is a large-scale
medical diagnosis JSON dataset from the NeurIPS 2022 Datasets and Benchmarks track.
The data was synthetically
generated from a validated medical knowledge base, producing deterministic
symp\-tom--diagnosis relationships, which explains the near-perfect
classification scores observed in the utility metric
(Table~\ref{tab:utility}) and makes the dataset valuable primarily as a
stress test for structural complexity and scale rather than
classification difficulty. Each of the 1{,}025{,}602 records contains
demographics, a pathology label (49 classes, our utility target), and two
variable-length arrays: symptom evidences and differential diagnoses with
associated probabilities. After flattening, the dataset expands to 100
columns with 67\% sparsity. At over one million records, DDXPlus is the
largest dataset in our benchmark.

\paragraph{GitHub Issues.}
This JSON-native dataset contains 642{,}099 public GitHub issue events
from GH Archive\footnote{Data obtained from hourly GH Archive dumps at
\url{https://data.gharchive.org/} for the week of 2026-02-23 through 2026-03-01.}.
Each record corresponds to an \texttt{IssuesEvent}, representing actions
such as opening, closing, labeling, or assigning an issue. We retain
structured issue metadata, including timestamps, state, lock and state
reasons, comment and reaction counts, user types, milestone presence,
and variable-length arrays of labels and assignees. We remove unique
identifiers, URLs, repository names, usernames, and free-text fields,
and compress label colors and names to avoid large quasi-free-text
fields while preserving the nested label-array structure. With
461 columns after flattening and 93\% sparsity, this dataset is the 
widest and sparsest in our evaluation. The utility target is the 
7-class \texttt{action} field.

\medskip
Table~\ref{tab:datasets} summarizes the dataset characteristics after
flattening, showing number of records, total number of columns, number of numeric columns, 
total count of discrete values across all columns and sparsity, which 
denotes the fraction of structurally absent cells (key not present in
the record), as opposed to explicit nulls.

\begin{table}[tb]
  \centering
  \small
  \caption{Dataset characteristics after flattening. Sparsity is measured before type separation as the fraction of structurally absent flattened key-path cells.}
  \label{tab:datasets}
  \begin{tabular}{lrrrrr}
    \toprule
    Dataset & Records & Total cols. & Num. & Discr. & Sparsity \\
    \midrule
    Adult & 48{,}842 & 15 & 6 & 104 & 0.0\% \\
    Diabetes & 81{,}413 & 37 & 13 & 2{,}261 & 0.0\% \\
    Electric Vehicles & 210{,}011 & 18 & 6 & 5{,}860 & 11.1\% \\
    Yelp & 150{,}346 & 142 & 6 & 26{,}812 & 77.8\% \\
    DDXPlus & 1{,}160{,}131 & 100 & 50 & 6{,}527 & 67.1\% \\
    GitHub Issues & 642{,}099 & 461 & 18 & 7{,}523 & 93.0\% \\
    \bottomrule
  \end{tabular}
\end{table}

 \subsection{Experiment Protocol}
\label{sec:protocol}

\paragraph{Data splits.}

For Adult and Diabetes, we use the exact preprocessing and train/test
splits provided by TabDiff~\cite{shiTabDiffMixedtypeDiffusion2025} to
ensure a direct comparison. For the remaining 4 datasets, we
split 80/10/10 into train, validation, and test sets. Hyperparameter
selection for \origami uses the inner train/validation split; final
evaluation uses the full training set and held-out test set. 
The privacy metrics evaluation requires different split proportions,
which is explained in Section~\ref{sec:privacy}.

\paragraph{Training.}
All training runs were conducted on a single NVIDIA V100 (16GB) GPU with a maximum wall-clock 
budget of 24 hours per run. Unless stated otherwise, we use the default hyperparameter 
configurations reported in each model's original publication. We made every effort to train 
each model on all datasets; for the largest 3 datasets (Yelp, DDXPlus, Github Issues), some models 
required reduced batch sizes and disabled costly validation-set evaluation during training.
Details for each model can be found in Section~\ref{sec:baselines}.

\paragraph{Sampling.}
For each trained model, we generate 3 replicate sample sets with 
different random seeds matching the size of the training split. Each sample set is evaluated individually
against fidelity, utility, detection and privacy metrics. We average
all scores and report mean and standard deviation across replicates 
in Tables~\ref{tab:fidelity}--\ref{tab:privacy}.

\subsection{Evaluation Metrics}
\label{sec:metrics}

We evaluate synthetic data along four dimensions: \emph{fidelity}, \emph{utility},
\emph{detection}, and \emph{privacy}. For easier comparison, all primary scores are normalized to $[0, 1]$ with 1.0
being best. Standard tabular evaluation metrics (e.g.,
SDMetrics~\cite{sdmetrics}) assume flat tables with single-type
columns and silently drop missing values. This makes them unsuitable
for semi-structured data where keys may be variably present,
heterogeneously typed, or structurally nested. We therefore extend the
standard metrics with modifications that explicitly account for
structural missingness and type polymorphism. All metrics operate on the
flattened, type-separated representation
(Section~\ref{sec:tabular-repr}), ensuring a consistent column schema
across synthesizers, including \origami. By design, our modified metrics
collapse to the standard metrics (e.g. provided by the SDMetrics package) 
for dense, homogeneous tables (Adult, Diabetes).

\paragraph{Fidelity.}
Fidelity measures how well synthetic data reproduces the statistical
properties of the training data. The overall fidelity score is the
average of two sub-scores: Column Shapes (marginal distributions) and
Column Pair Trends (pairwise dependencies).

For \emph{Column Shapes}, standard practice computes a single
similarity statistic per column---Kolmogorov--Smirnov (KS) Complement
for numeric columns, Total Variation (TV) Complement for categorical
columns---and averages across columns.
This fails on type-separated data: a column like
\texttt{awards.wins.num} may be \texttt{NaN} in rows where the key
is absent or holds a non-numeric type, and simply dropping these rows
conflates structural absence with distributional similarity.

We instead decompose the per-column joint distribution over presence,
type, and value via the chain rule, and measure real-vs-synthetic
similarity at each factor independently for every column:
\begin{align}
  \varphi_{\text{pres}}(c) &= \operatorname{sim}\bigl(
    P_r(\text{Pres} \mid c),\;
    P_s(\text{Pres} \mid c)\bigr) \\
  \varphi_{\text{type}}(c) &= \operatorname{sim}\bigl(
    P_r(T \mid c, \text{pres}),\;
    P_s(T \mid c, \text{pres})\bigr) \\
  \varphi_{\text{val}}(c) &= \operatorname{sim}\bigl(
    P_r(V \mid c, \text{pres}, T),\;
    P_s(V \mid c, \text{pres}, T)\bigr)
\end{align}
where $P_r$ and $P_s$ denote the real and synthetic distributions for given column $c$,
$\operatorname{sim}$ is TV Complement for discrete distributions and
KS Complement for continuous ones, and each factor conditions on the
previous level. The per-column fidelity score $\varphi(c)$ combines these
multiplicatively:
\begin{equation}
  \varphi(c) = \varphi_{\text{pres}}(c) \cdot
               \varphi_{\text{type}}(c) \cdot
               \varphi_{\text{val}}(c)
\end{equation}

For \emph{Column Pair Trends}, we compute pairwise similarity on all
type-separated columns, including type indicator and split sub-type
columns when multiple types are present. Continuous--continuous pairs use Correlation
Similarity ($1 - |r_{\text{real}} - r_{\text{synth}}| / 2$),
discrete--discrete pairs use 2D TV Complement on the joint distribution,
and mixed pairs discretize the continuous column into 10 bins, following
the widely used SDMetrics implementation. Pairs are
weighted by co-occurrence rate
$w(a,b) \propto \max(\text{co\_rate}_{\text{real}},
\text{co\_rate}_{\text{synth}})$, naturally down-weighting pairs that
rarely co-occur due to structural missingness.

\paragraph{Utility.}
ML utility (sometimes referred to as ML efficacy) measures whether synthetic data preserves the predictive
structure of the original data. We follow the Train-Syn\-thetic-Test-Real
(TSTR) protocol~\cite{yuan_multi-faceted_2025}: train a classifier on synthetic data and
evaluate on held-out real data, then compare against a baseline model
trained on real data (TRTR). All datasets in our benchmark are
classification tasks. The utility score is the ratio of TSTR to TRTR
weighted $F_1$, capped at 1.0:
\begin{equation}
  \text{Utility} = \min\!\left(
    \frac{F_{1}^{\text{TSTR}}}{F_{1}^{\text{TRTR}}},\; 1.0
  \right)
\end{equation}
We report the TSTR $F_1$ scores alongside the ratio to
allow direct comparison. XGBoost is configured with default
hyperparameters (\texttt{n\_estimators}=100, \texttt{max\_depth}=6)
and native categorical support to avoid one-hot explosion on
high-cardinality features.

\paragraph{Detection.}
Detection measures whether a classifier can distinguish synthetic
records from real ones, following the Classifier Two-Sample Test
(C2ST) framework~\cite{lopez-pazRevisitingClassifierTwoSample2018}. 
The standard approach in the
tabular synthesis literature uses logistic regression for
this test~\cite{sdmetrics}, but we observe that the default
solver frequently fails to converge within the iteration budget for large datasets,
producing an undertrained classifier that cannot reliably separate
real from synthetic data. This artificially inflates detection scores.
Consequently, we replace logistic regression with a small, shallow 
XGBoost classifier (\texttt{n\_estimators}=10, \texttt{max\_depth}=3, native
categorical support), which converges reliably and can detect
non-linear feature interactions that a linear classifier like logistic
regression cannot separate. Stratified 3-fold cross-validation is used
to avoid overfitting. Per-fold ROC AUC is clamped to $[0.5, 1.0]$
and transformed:
\begin{equation}
  \text{Detection} = 1 - \text{mean}\bigl(
    \max(0.5,\; \text{AUC}_j) \cdot 2 - 1
  \bigr)
\end{equation}
where $\text{AUC}_j$ is the ROC AUC on fold $j$ of a stratified $k$-fold cross-validation.
A score of 1.0 means the classifier performs at chance
(indistinguishable); 0.0 means perfect separation. We report the
raw (unclamped) ROC AUC alongside the detection score for more
granular comparison.

\paragraph{Privacy.}
\label{sec:privacy}
Privacy measures whether the synthesizer memorizes training records. We adopt the Distance to Closest Record (DCR)
methodology~\cite{liuScalingPrivacyPreserving2024}.

For privacy evaluation, we require equally sized reference sets to
avoid biasing the DCR toward the larger set. We therefore create a
separate 50/50 split from the original training partition, train a
new model on one half, and generate an equal number of synthetic
records. This yields equally sized train, test, and synthetic sets
for distance computation.

For each synthetic record, we compute
its $L_2$ distance to the nearest training record and the nearest test
record in the one-hot encoded, range-normalized feature space. If the
synthetic data does not memorize training data, it should be equally
likely to be closer to either reference set:
\begin{equation}
  \text{DCR} = \frac{|\{s \in S : d(s, \text{Train}) < d(s, \text{Test})\}|}
               {|S|} \cdot 100\%
\end{equation}
A DCR of 50\% is ideal. The privacy score penalizes only values
above 50\% (closer to train = memorization):
\begin{equation}
\text{Privacy} = 1 - 2 \cdot \max(\text{DCR}/100 - 0.5,\; 0)
\end{equation}

We additionally record the number of exact matches (zero-distance pairs) between synthetic
and training records as a direct memorization indicator, and discuss the cases where such
matches occur in Section~\ref{sec:main-results}.

\subsection{Model Training Details}
\label{sec:baselines}

We compare \origami against six baselines covering all different architecture families: 
TVAE (VAE), CTGAN (GAN), REaLTabFormer and TabularARGN (autoregressive), Tabby (LLM), TabDiff (diffusion). 
We additionally attempted to include GReaT~\cite{borisovLanguageModelsAre2023} 
but were unable to obtain results due to sampling failures and architectural limitations, as discussed below.

\paragraph{TVAE and CTGAN} 
We use the SDV \cite{patkiSyntheticDataVault2016} library implementation with default
hyperparameters (300 epochs) for both models. Both TVAE and CTGAN exceeded the maximum available memory for the Electric Vehicles, Yelp,  DDXPlus and Github Issues datasets due to their one-hot encoding of high-cardinality categorical columns.\footnote{The SDV library itself warns about this, e.g.: \emph{``PerformanceAlert: Using the CTGANSynthesizer on this data is not recommended''}. }
We marked the entries for these datasets OOM in the results tables. 

\paragraph{REalTabFormer}
We use the published implementation with default settings. For the Yelp
dataset, we had to disable the sensitivity-based early stopping
criterion (setting \texttt{n\_critic=0}) due to memory limitations and instead
rely on the built-in fallback mechanism using loss convergence. On 
Electric Vehicles, training was interrupted after 24h at epoch 40. 
REalTabFormer was unable to train within the memory budget on DDXPlus
and Github Issues, even without sensitivity analysis and with reduced batch size 
(marked OOM in the results table).

\paragraph{TabDiff}
For Adult, Diabetes, and Electric Vehicles, we train for the default 8{,}000 epochs
with the default batch size of 4{,}096 and learning rate of $10^{-3}$,
following the original paper. For Yelp, DDXPlus and Github Issues, we had to reduce the batch
size to 512 to fit in GPU memory. For those runs, we also adjusted the learning 
rate to $3.5 \cdot 10^{-4}$ following square-root scaling practices. We also observed loss spikes during training
and added gradient clipping at 1.0 to stabilize convergence. We further
needed to disable evaluation during training due to the computational cost of sampling
(e.g. several hours for a sample of DDXPlus). Tabdiff reached 461, 392 and 33 epochs on
Yelp, DDXplus and Github Issues respectively before exceeding the 24h training budget. We 
sampled from and evaluated the best checkpoint by validation loss. 

\paragraph{TabularARGN}
TabularARGN is available through the freely available \texttt{mostlyai-engine}\footnote{\url{https://github.com/mostly-ai/mostlyai-engine}} 
Python package, released by Mostly AI, a synthetic
data generation company. We use the package's default configuration.
Training completed on Adult, Diabetes, Electric Vehicles, and Yelp. For
DDXPlus and GitHub Issues, training reached the 24-hour wall-clock
budget after 61 and 25 epochs, respectively; we evaluate the best
checkpoint by validation loss.

\paragraph{GReaT and Tabby}
Both GReaT~\cite{borisovLanguageModelsAre2023} and its Mixture-of-Experts extension 
Tabby~\cite{crompTabbyLanguageModel2026} build on GPT-2 backbones with a fixed context 
window of 1{,}024 tokens. For Diabetes, Yelp, DDXPlus and Github Issues, flattened rows 
exceed this limit, making these datasets incompatible with GPT-2 family architectures. 
On Adult, GReaT trained successfully but its sampling procedure repeatedly failed to 
produce valid rows; we therefore excluded it from the results. Tabby produced results
on Adult only; all other datasets are marked OOM.

\paragraph{\origami}
We use the same transformer backbone family across all datasets: 8
layers, 4 attention heads, feed-forward dimension 512, 
key-order shuffling, and grammar/schema constraints during training and
inference. Model width and numeric heads are scaled modestly with dataset
complexity: $d_{\mathrm{model}}=64$ for Adult, Diabetes, Yelp, and
DDXPlus, and $128$ for Electric Vehicles and GitHub Issues. The
continuous head uses 5 mixture components except for GitHub Issues,
where we use 10. Batch size is 128 for the tabular datasets and 64 for
the JSON-native datasets. Learning rates range from $10^{-4}$ to
$5\cdot10^{-4}$ under cosine decay, and training runs for 20--400 epochs
depending on dataset size, evaluating on the final checkpoint.
All exact hyperparameters are included in the released configuration
files.

%% file: 05_discussion.tex
\section{Discussion}
\label{sec:discussion}

\subsection{Main Results}
\label{sec:main-results}

\begin{table}[tb]
  \centering
  \caption{Training cost comparison on the Yelp dataset.}
  \label{tab:training-cost}
  \begin{tabular}{lrrrr}
    \toprule
    \textbf{Model} & \textbf{Params} & \textbf{Time/Epoch} & \textbf{Epochs} & \textbf{Total} \\
    \midrule
    REalTabFormer  & 59.4M & 2,160 sec & 40 & $>$24h \\
    TabDiff        & 25.8M & 187 sec & 461  & $>$24h \\
    TabularARGN    & 9.0M  & 295 sec & 77  & 6.3h  \\
    Origami        & 1.7M & 239 sec & 200 & 13.3h \\
    \bottomrule
  \end{tabular}
\end{table}

\begin{table*}[t]
  \caption{Fidelity metrics across datasets (mean $\pm$ std over 3 replicates). Higher is better.}
  \label{tab:fidelity}
  \small
  \begin{tabular}{llccccccc}
    \toprule
     &  & Tabby & TVAE & CTGAN & REaLTabFormer & TabularARGN & TabDiff & \origami (ours) \\
    \midrule
    \multirow{3}{*}{Adult} &   Overall score & $0.938{\color{gray}\scriptstyle\,\pm\,0.006}$ & $0.893{\color{gray}\scriptstyle\,\pm\,0.001}$ & $0.876{\color{gray}\scriptstyle\,\pm\,0.000}$ & $0.960{\color{gray}\scriptstyle\,\pm\,0.000}$ & $0.979{\color{gray}\scriptstyle\,\pm\,0.000}$ & $0.990{\color{gray}\scriptstyle\,\pm\,0.000}$ & $\underline{\mathbf{0.992}}{\color{gray}\scriptstyle\,\pm\,0.000}$ \\
     &   \quad Shapes & $0.946{\color{gray}\scriptstyle\,\pm\,0.000}$ & $0.905{\color{gray}\scriptstyle\,\pm\,0.001}$ & $0.871{\color{gray}\scriptstyle\,\pm\,0.001}$ & $0.967{\color{gray}\scriptstyle\,\pm\,0.001}$ & $0.985{\color{gray}\scriptstyle\,\pm\,0.000}$ & $0.994{\color{gray}\scriptstyle\,\pm\,0.000}$ & $0.996{\color{gray}\scriptstyle\,\pm\,0.000}$ \\
     &   \quad Trends & $0.930{\color{gray}\scriptstyle\,\pm\,0.011}$ & $0.881{\color{gray}\scriptstyle\,\pm\,0.001}$ & $0.882{\color{gray}\scriptstyle\,\pm\,0.000}$ & $0.954{\color{gray}\scriptstyle\,\pm\,0.001}$ & $0.973{\color{gray}\scriptstyle\,\pm\,0.001}$ & $0.985{\color{gray}\scriptstyle\,\pm\,0.001}$ & $0.989{\color{gray}\scriptstyle\,\pm\,0.000}$ \\
    \addlinespace
    \multirow{3}{*}{Diabetes} &   Overall score & OOM & $0.860{\color{gray}\scriptstyle\,\pm\,0.000}$ & $0.927{\color{gray}\scriptstyle\,\pm\,0.000}$ & $0.965{\color{gray}\scriptstyle\,\pm\,0.000}$ & $0.982{\color{gray}\scriptstyle\,\pm\,0.000}$ & $0.982{\color{gray}\scriptstyle\,\pm\,0.000}$ & $\underline{\mathbf{0.992}}{\color{gray}\scriptstyle\,\pm\,0.000}$ \\
     &   \quad Shapes & -- & $0.890{\color{gray}\scriptstyle\,\pm\,0.000}$ & $0.943{\color{gray}\scriptstyle\,\pm\,0.000}$ & $0.973{\color{gray}\scriptstyle\,\pm\,0.000}$ & $0.988{\color{gray}\scriptstyle\,\pm\,0.000}$ & $0.988{\color{gray}\scriptstyle\,\pm\,0.000}$ & $0.996{\color{gray}\scriptstyle\,\pm\,0.000}$ \\
     &   \quad Trends & -- & $0.830{\color{gray}\scriptstyle\,\pm\,0.000}$ & $0.911{\color{gray}\scriptstyle\,\pm\,0.000}$ & $0.957{\color{gray}\scriptstyle\,\pm\,0.001}$ & $0.976{\color{gray}\scriptstyle\,\pm\,0.000}$ & $0.975{\color{gray}\scriptstyle\,\pm\,0.000}$ & $0.989{\color{gray}\scriptstyle\,\pm\,0.000}$ \\
    \addlinespace
    \multirow{3}{*}{\shortstack{Electric\\Vehicles}} &   Overall score & OOM & OOM & OOM & $0.866{\color{gray}\scriptstyle\,\pm\,0.000}$ & $0.968{\color{gray}\scriptstyle\,\pm\,0.000}$ & $0.976{\color{gray}\scriptstyle\,\pm\,0.000}$ & $\underline{\mathbf{0.987}}{\color{gray}\scriptstyle\,\pm\,0.000}$ \\
     &   \quad Shapes & -- & -- & -- & $0.896{\color{gray}\scriptstyle\,\pm\,0.000}$ & $0.973{\color{gray}\scriptstyle\,\pm\,0.000}$ & $0.983{\color{gray}\scriptstyle\,\pm\,0.000}$ & $0.993{\color{gray}\scriptstyle\,\pm\,0.000}$ \\
     &   \quad Trends & -- & -- & -- & $0.836{\color{gray}\scriptstyle\,\pm\,0.000}$ & $0.963{\color{gray}\scriptstyle\,\pm\,0.000}$ & $0.969{\color{gray}\scriptstyle\,\pm\,0.000}$ & $0.982{\color{gray}\scriptstyle\,\pm\,0.000}$ \\
    \addlinespace
    \multirow{3}{*}{Yelp} &   Overall score & OOM & OOM & OOM & $0.895{\color{gray}\scriptstyle\,\pm\,0.000}$ & $0.883{\color{gray}\scriptstyle\,\pm\,0.000}$ & $0.914{\color{gray}\scriptstyle\,\pm\,0.000}$ & $\underline{\mathbf{0.960}}{\color{gray}\scriptstyle\,\pm\,0.000}$ \\
     &   \quad Shapes & -- & -- & -- & $0.878{\color{gray}\scriptstyle\,\pm\,0.001}$ & $0.871{\color{gray}\scriptstyle\,\pm\,0.001}$ & $0.910{\color{gray}\scriptstyle\,\pm\,0.000}$ & $0.966{\color{gray}\scriptstyle\,\pm\,0.000}$ \\
     &   \quad Trends & -- & -- & -- & $0.911{\color{gray}\scriptstyle\,\pm\,0.000}$ & $0.894{\color{gray}\scriptstyle\,\pm\,0.000}$ & $0.918{\color{gray}\scriptstyle\,\pm\,0.000}$ & $0.955{\color{gray}\scriptstyle\,\pm\,0.000}$ \\
    \addlinespace
    \multirow{3}{*}{DDXPlus} &   Overall score & OOM & OOM & OOM & OOM & $0.797{\color{gray}\scriptstyle\,\pm\,0.000}$ & $0.857{\color{gray}\scriptstyle\,\pm\,0.000}$ & $\underline{\mathbf{0.917}}{\color{gray}\scriptstyle\,\pm\,0.000}$ \\
     &   \quad Shapes & -- & -- & -- & -- & $0.792{\color{gray}\scriptstyle\,\pm\,0.000}$ & $0.846{\color{gray}\scriptstyle\,\pm\,0.000}$ & $0.930{\color{gray}\scriptstyle\,\pm\,0.000}$ \\
     &   \quad Trends & -- & -- & -- & -- & $0.803{\color{gray}\scriptstyle\,\pm\,0.000}$ & $0.868{\color{gray}\scriptstyle\,\pm\,0.000}$ & $0.905{\color{gray}\scriptstyle\,\pm\,0.000}$ \\
    \addlinespace
    \multirow{3}{*}{\shortstack{GitHub\\Issues}} &   Overall score & OOM & OOM & OOM & OOM & $0.910{\color{gray}\scriptstyle\,\pm\,0.001}$ & $0.904{\color{gray}\scriptstyle\,\pm\,0.027}$ & $\underline{\mathbf{0.930}}{\color{gray}\scriptstyle\,\pm\,0.000}$ \\
     &   \quad Shapes & -- & -- & -- & -- & $0.958{\color{gray}\scriptstyle\,\pm\,0.000}$ & $0.932{\color{gray}\scriptstyle\,\pm\,0.002}$ & $0.966{\color{gray}\scriptstyle\,\pm\,0.000}$ \\
     &   \quad Trends & -- & -- & -- & -- & $0.861{\color{gray}\scriptstyle\,\pm\,0.002}$ & $0.875{\color{gray}\scriptstyle\,\pm\,0.053}$ & $0.894{\color{gray}\scriptstyle\,\pm\,0.000}$ \\
    \bottomrule
  \end{tabular}
\end{table*}

\begin{table*}[t]
  \caption{Utility metrics across datasets (mean $\pm$ std over 3 replicates). Higher is better. Overall utility normalizes TSTR $F_1$ by the corresponding real-data baseline.}
  \label{tab:utility}
  \small
  \begin{tabular}{llccccccc}
    \toprule
     &  & Tabby & TVAE & CTGAN & REaLTabFormer & TabularARGN & TabDiff & \origami (ours) \\
    \midrule
    \multirow{2}{*}{Adult} &   Overall score & $0.948{\color{gray}\scriptstyle\,\pm\,0.002}$ & $0.950{\color{gray}\scriptstyle\,\pm\,0.002}$ & $0.893{\color{gray}\scriptstyle\,\pm\,0.004}$ & $0.994{\color{gray}\scriptstyle\,\pm\,0.001}$ & $0.983{\color{gray}\scriptstyle\,\pm\,0.002}$ & $0.982{\color{gray}\scriptstyle\,\pm\,0.002}$ & $\underline{\mathbf{0.994}}{\color{gray}\scriptstyle\,\pm\,0.002}$ \\
     &   \quad TSTR $F_1$ & $0.821{\color{gray}\scriptstyle\,\pm\,0.002}$ & $0.822{\color{gray}\scriptstyle\,\pm\,0.002}$ & $0.773{\color{gray}\scriptstyle\,\pm\,0.003}$ & $0.860{\color{gray}\scriptstyle\,\pm\,0.000}$ & $0.851{\color{gray}\scriptstyle\,\pm\,0.002}$ & $0.850{\color{gray}\scriptstyle\,\pm\,0.001}$ & $0.861{\color{gray}\scriptstyle\,\pm\,0.002}$ \\
    \addlinespace
    \multirow{2}{*}{Diabetes} &   Overall score & OOM & $0.954{\color{gray}\scriptstyle\,\pm\,0.011}$ & $0.926{\color{gray}\scriptstyle\,\pm\,0.003}$ & $0.973{\color{gray}\scriptstyle\,\pm\,0.003}$ & $0.973{\color{gray}\scriptstyle\,\pm\,0.006}$ & $0.976{\color{gray}\scriptstyle\,\pm\,0.006}$ & $\underline{\mathbf{0.988}}{\color{gray}\scriptstyle\,\pm\,0.002}$ \\
     &   \quad TSTR $F_1$ & -- & $0.592{\color{gray}\scriptstyle\,\pm\,0.007}$ & $0.574{\color{gray}\scriptstyle\,\pm\,0.002}$ & $0.603{\color{gray}\scriptstyle\,\pm\,0.002}$ & $0.602{\color{gray}\scriptstyle\,\pm\,0.004}$ & $0.605{\color{gray}\scriptstyle\,\pm\,0.004}$ & $0.612{\color{gray}\scriptstyle\,\pm\,0.001}$ \\
    \addlinespace
    \multirow{2}{*}{\shortstack{Electric\\Vehicles}} &   Overall score & OOM & OOM & OOM & $0.955{\color{gray}\scriptstyle\,\pm\,0.009}$ & $0.983{\color{gray}\scriptstyle\,\pm\,0.004}$ & $0.978{\color{gray}\scriptstyle\,\pm\,0.003}$ & $\underline{\mathbf{0.999}}{\color{gray}\scriptstyle\,\pm\,0.001}$ \\
     &   \quad TSTR $F_1$ & -- & -- & -- & $0.923{\color{gray}\scriptstyle\,\pm\,0.009}$ & $0.950{\color{gray}\scriptstyle\,\pm\,0.004}$ & $0.945{\color{gray}\scriptstyle\,\pm\,0.003}$ & $0.966{\color{gray}\scriptstyle\,\pm\,0.001}$ \\
    \addlinespace
    \multirow{2}{*}{Yelp} &   Overall score & OOM & OOM & OOM & $\underline{\mathbf{0.981}}{\color{gray}\scriptstyle\,\pm\,0.002}$ & $0.971{\color{gray}\scriptstyle\,\pm\,0.003}$ & $0.947{\color{gray}\scriptstyle\,\pm\,0.002}$ & $0.978{\color{gray}\scriptstyle\,\pm\,0.002}$ \\
     &   \quad TSTR $F_1$ & -- & -- & -- & $0.828{\color{gray}\scriptstyle\,\pm\,0.001}$ & $0.820{\color{gray}\scriptstyle\,\pm\,0.002}$ & $0.800{\color{gray}\scriptstyle\,\pm\,0.002}$ & $0.826{\color{gray}\scriptstyle\,\pm\,0.002}$ \\
    \addlinespace
    \multirow{2}{*}{DDXPlus} &   Overall score & OOM & OOM & OOM & OOM & $0.998{\color{gray}\scriptstyle\,\pm\,0.003}$ & $\underline{\mathbf{1.000}}{\color{gray}\scriptstyle\,\pm\,0.000}$ & $\underline{\mathbf{1.000}}{\color{gray}\scriptstyle\,\pm\,0.000}$ \\
     &   \quad TSTR $F_1$ & -- & -- & -- & -- & $0.981{\color{gray}\scriptstyle\,\pm\,0.005}$ & $0.987{\color{gray}\scriptstyle\,\pm\,0.001}$ & $0.995{\color{gray}\scriptstyle\,\pm\,0.000}$ \\
    \addlinespace
    \multirow{2}{*}{\shortstack{GitHub\\Issues}} &   Overall score & OOM & OOM & OOM & OOM & $0.961{\color{gray}\scriptstyle\,\pm\,0.001}$ & $0.949{\color{gray}\scriptstyle\,\pm\,0.003}$ & $\underline{\mathbf{0.979}}{\color{gray}\scriptstyle\,\pm\,0.001}$ \\
     &   \quad TSTR $F_1$ & -- & -- & -- & -- & $0.680{\color{gray}\scriptstyle\,\pm\,0.001}$ & $0.672{\color{gray}\scriptstyle\,\pm\,0.001}$ & $0.693{\color{gray}\scriptstyle\,\pm\,0.000}$ \\
    \bottomrule
  \end{tabular}
\end{table*}

\begin{table*}[t]
  \caption{Detection metrics across datasets (mean $\pm$ std over 3 replicates). Detection score: higher means harder to detect (better). XGBoost classifier ROC AUC: lower means harder to distinguish from real data (better).}
  \label{tab:detection}
  \small
  \begin{tabular}{llccccccc}
    \toprule
     &  & Tabby & TVAE & CTGAN & REaLTabFormer & TabularARGN & TabDiff & \origami (ours) \\
    \midrule
    \multirow{2}{*}{Adult} &   Overall score $\uparrow$ & $0.587{\color{gray}\scriptstyle\,\pm\,0.006}$ & $0.218{\color{gray}\scriptstyle\,\pm\,0.004}$ & $0.112{\color{gray}\scriptstyle\,\pm\,0.001}$ & $0.807{\color{gray}\scriptstyle\,\pm\,0.007}$ & $0.866{\color{gray}\scriptstyle\,\pm\,0.006}$ & $0.967{\color{gray}\scriptstyle\,\pm\,0.001}$ & $\underline{\mathbf{0.979}}{\color{gray}\scriptstyle\,\pm\,0.002}$ \\
     &   \quad ROC AUC $\downarrow$ & $0.707{\color{gray}\scriptstyle\,\pm\,0.003}$ & $0.891{\color{gray}\scriptstyle\,\pm\,0.002}$ & $0.944{\color{gray}\scriptstyle\,\pm\,0.000}$ & $0.596{\color{gray}\scriptstyle\,\pm\,0.004}$ & $0.567{\color{gray}\scriptstyle\,\pm\,0.003}$ & $0.517{\color{gray}\scriptstyle\,\pm\,0.000}$ & $0.511{\color{gray}\scriptstyle\,\pm\,0.001}$ \\
    \addlinespace
    \multirow{2}{*}{Diabetes} &   Overall score $\uparrow$ & OOM & $0.045{\color{gray}\scriptstyle\,\pm\,0.001}$ & $0.411{\color{gray}\scriptstyle\,\pm\,0.004}$ & $0.696{\color{gray}\scriptstyle\,\pm\,0.001}$ & $0.896{\color{gray}\scriptstyle\,\pm\,0.003}$ & $0.885{\color{gray}\scriptstyle\,\pm\,0.003}$ & $\underline{\mathbf{1.000}}{\color{gray}\scriptstyle\,\pm\,0.000}$ \\
     &   \quad ROC AUC $\downarrow$ & -- & $0.978{\color{gray}\scriptstyle\,\pm\,0.000}$ & $0.794{\color{gray}\scriptstyle\,\pm\,0.002}$ & $0.652{\color{gray}\scriptstyle\,\pm\,0.001}$ & $0.552{\color{gray}\scriptstyle\,\pm\,0.001}$ & $0.557{\color{gray}\scriptstyle\,\pm\,0.001}$ & $0.485{\color{gray}\scriptstyle\,\pm\,0.001}$ \\
    \addlinespace
    \multirow{2}{*}{\shortstack{Electric\\Vehicles}} &   Overall score $\uparrow$ & OOM & OOM & OOM & $0.417{\color{gray}\scriptstyle\,\pm\,0.001}$ & $0.640{\color{gray}\scriptstyle\,\pm\,0.009}$ & $0.937{\color{gray}\scriptstyle\,\pm\,0.003}$ & $\underline{\mathbf{1.000}}{\color{gray}\scriptstyle\,\pm\,0.000}$ \\
     &   \quad ROC AUC $\downarrow$ & -- & -- & -- & $0.791{\color{gray}\scriptstyle\,\pm\,0.001}$ & $0.680{\color{gray}\scriptstyle\,\pm\,0.004}$ & $0.531{\color{gray}\scriptstyle\,\pm\,0.001}$ & $0.497{\color{gray}\scriptstyle\,\pm\,0.002}$ \\
    \addlinespace
    \multirow{2}{*}{Yelp} &   Overall score $\uparrow$ & OOM & OOM & OOM & $0.327{\color{gray}\scriptstyle\,\pm\,0.001}$ & $0.341{\color{gray}\scriptstyle\,\pm\,0.006}$ & $0.427{\color{gray}\scriptstyle\,\pm\,0.010}$ & $\underline{\mathbf{0.772}}{\color{gray}\scriptstyle\,\pm\,0.003}$ \\
     &   \quad ROC AUC $\downarrow$ & -- & -- & -- & $0.837{\color{gray}\scriptstyle\,\pm\,0.000}$ & $0.829{\color{gray}\scriptstyle\,\pm\,0.003}$ & $0.787{\color{gray}\scriptstyle\,\pm\,0.005}$ & $0.614{\color{gray}\scriptstyle\,\pm\,0.002}$ \\
    \addlinespace
    \multirow{2}{*}{DDXPlus} &   Overall score $\uparrow$ & OOM & OOM & OOM & OOM & $0.400{\color{gray}\scriptstyle\,\pm\,0.002}$ & $0.133{\color{gray}\scriptstyle\,\pm\,0.010}$ & $\underline{\mathbf{0.558}}{\color{gray}\scriptstyle\,\pm\,0.013}$ \\
     &   \quad ROC AUC $\downarrow$ & -- & -- & -- & -- & $0.800{\color{gray}\scriptstyle\,\pm\,0.001}$ & $0.934{\color{gray}\scriptstyle\,\pm\,0.005}$ & $0.721{\color{gray}\scriptstyle\,\pm\,0.007}$ \\
    \addlinespace
    \multirow{2}{*}{\shortstack{GitHub\\Issues}} &   Overall score $\uparrow$ & OOM & OOM & OOM & OOM & $0.676{\color{gray}\scriptstyle\,\pm\,0.008}$ & $0.449{\color{gray}\scriptstyle\,\pm\,0.011}$ & $\underline{\mathbf{0.687}}{\color{gray}\scriptstyle\,\pm\,0.003}$ \\
     &   \quad ROC AUC $\downarrow$ & -- & -- & -- & -- & $0.662{\color{gray}\scriptstyle\,\pm\,0.004}$ & $0.775{\color{gray}\scriptstyle\,\pm\,0.005}$ & $0.656{\color{gray}\scriptstyle\,\pm\,0.001}$ \\
    \bottomrule
  \end{tabular}
\end{table*}

\begin{table*}[t]
  \caption{Privacy metrics across datasets (mean $\pm$ std over 3 replicates). Privacy score: higher is better. DCR score $\leq$ 50 indicates no memorization. Exact-match counts are discussed in the text.}
  \label{tab:privacy}
  \small
  \begin{tabular}{llccccccc}
    \toprule
     &  & Tabby & TVAE & CTGAN & REaLTabFormer & TabularARGN & TabDiff & \origami (ours) \\
    \midrule
    \multirow{2}{*}{Adult} &   Overall score $\uparrow$ & $\underline{\mathbf{1.000}}{\color{gray}\scriptstyle\,\pm\,0.000}$ & $0.967{\color{gray}\scriptstyle\,\pm\,0.006}$ & $\underline{\mathbf{1.000}}{\color{gray}\scriptstyle\,\pm\,0.000}$ & $0.915{\color{gray}\scriptstyle\,\pm\,0.004}$ & $0.981{\color{gray}\scriptstyle\,\pm\,0.006}$ & $0.831{\color{gray}\scriptstyle\,\pm\,0.001}$ & $0.984{\color{gray}\scriptstyle\,\pm\,0.000}$ \\
     &   \quad DCR $\downarrow$ & $49.204{\color{gray}\scriptstyle\,\pm\,0.420}$ & $51.663{\color{gray}\scriptstyle\,\pm\,0.321}$ & $49.648{\color{gray}\scriptstyle\,\pm\,0.276}$ & $54.231{\color{gray}\scriptstyle\,\pm\,0.182}$ & $50.938{\color{gray}\scriptstyle\,\pm\,0.314}$ & $58.471{\color{gray}\scriptstyle\,\pm\,0.070}$ & $50.805{\color{gray}\scriptstyle\,\pm\,0.025}$ \\
    \addlinespace
    \multirow{2}{*}{Diabetes} &   Overall score $\uparrow$ & OOM & $0.967{\color{gray}\scriptstyle\,\pm\,0.009}$ & $\underline{\mathbf{1.000}}{\color{gray}\scriptstyle\,\pm\,0.000}$ & $0.874{\color{gray}\scriptstyle\,\pm\,0.001}$ & $0.963{\color{gray}\scriptstyle\,\pm\,0.003}$ & $0.941{\color{gray}\scriptstyle\,\pm\,0.006}$ & $0.976{\color{gray}\scriptstyle\,\pm\,0.001}$ \\
     &   \quad DCR $\downarrow$ & -- & $51.643{\color{gray}\scriptstyle\,\pm\,0.456}$ & $49.504{\color{gray}\scriptstyle\,\pm\,0.151}$ & $56.293{\color{gray}\scriptstyle\,\pm\,0.068}$ & $51.869{\color{gray}\scriptstyle\,\pm\,0.154}$ & $52.936{\color{gray}\scriptstyle\,\pm\,0.310}$ & $51.224{\color{gray}\scriptstyle\,\pm\,0.054}$ \\
    \addlinespace
    \multirow{2}{*}{\shortstack{Electric\\Vehicles}} &   Overall score $\uparrow$ & OOM & OOM & OOM & $0.161{\color{gray}\scriptstyle\,\pm\,0.001}$ & $0.958{\color{gray}\scriptstyle\,\pm\,0.005}$ & $0.938{\color{gray}\scriptstyle\,\pm\,0.001}$ & $\underline{\mathbf{0.965}}{\color{gray}\scriptstyle\,\pm\,0.001}$ \\
     &   \quad DCR $\downarrow$ & -- & -- & -- & $91.951{\color{gray}\scriptstyle\,\pm\,0.071}$ & $52.081{\color{gray}\scriptstyle\,\pm\,0.273}$ & $53.125{\color{gray}\scriptstyle\,\pm\,0.056}$ & $51.770{\color{gray}\scriptstyle\,\pm\,0.072}$ \\
    \addlinespace
    \multirow{2}{*}{Yelp} &   Overall score $\uparrow$ & OOM & OOM & OOM & $0.939{\color{gray}\scriptstyle\,\pm\,0.002}$ & $0.955{\color{gray}\scriptstyle\,\pm\,0.003}$ & $0.971{\color{gray}\scriptstyle\,\pm\,0.004}$ & $\underline{\mathbf{0.974}}{\color{gray}\scriptstyle\,\pm\,0.001}$ \\
     &   \quad DCR $\downarrow$ & -- & -- & -- & $53.029{\color{gray}\scriptstyle\,\pm\,0.118}$ & $52.266{\color{gray}\scriptstyle\,\pm\,0.168}$ & $51.433{\color{gray}\scriptstyle\,\pm\,0.187}$ & $51.298{\color{gray}\scriptstyle\,\pm\,0.074}$ \\
    \addlinespace
    \multirow{2}{*}{DDXPlus} &   Overall score $\uparrow$ & OOM & OOM & OOM & OOM & $0.978{\color{gray}\scriptstyle\,\pm\,0.001}$ & $0.977{\color{gray}\scriptstyle\,\pm\,0.001}$ & $\underline{\mathbf{0.995}}{\color{gray}\scriptstyle\,\pm\,0.000}$ \\
     &   \quad DCR $\downarrow$ & -- & -- & -- & -- & $51.086{\color{gray}\scriptstyle\,\pm\,0.028}$ & $51.134{\color{gray}\scriptstyle\,\pm\,0.029}$ & $50.227{\color{gray}\scriptstyle\,\pm\,0.009}$ \\
    \addlinespace
    \multirow{2}{*}{\shortstack{GitHub\\Issues}} &   Overall score $\uparrow$ & OOM & OOM & OOM & OOM & $\underline{\mathbf{0.999}}{\color{gray}\scriptstyle\,\pm\,0.001}$ & $0.999{\color{gray}\scriptstyle\,\pm\,0.001}$ & $0.994{\color{gray}\scriptstyle\,\pm\,0.000}$ \\
     &   \quad DCR $\downarrow$ & -- & -- & -- & -- & $50.013{\color{gray}\scriptstyle\,\pm\,0.062}$ & $50.065{\color{gray}\scriptstyle\,\pm\,0.067}$ & $50.288{\color{gray}\scriptstyle\,\pm\,0.010}$ \\
    \bottomrule
  \end{tabular}
\end{table*}

Tables~\ref{tab:fidelity}--\ref{tab:privacy} summarize the evaluation results averaged over 3 replicate sample sets, with the best
scores underlined per row. 
\origami achieves the highest fidelity and detection scores across all six datasets. Utility is best or tied-best on 
five of six datasets, with REaLTabFormer slightly ahead on Yelp (0.981 vs.\ 0.978). On the dense benchmarks (Adult,
Diabetes), \origami{}'s margins over the best baselines are small: the average fidelity gain is 0.006, and all 
competitive methods score above 0.96 on fidelity and utility. This indicates that these smaller tabular datasets
are reaching saturation as benchmarks for modern mixed-type synthesizers. 

As sparsity and structure increase, the gap widens. On the JSON-native datasets, \origami improves fidelity
over the best baseline by 0.046 on Yelp, 0.060 on DDXPlus, and 0.020 on GitHub Issues. Detection gains are also larger, 
especially on Yelp (+0.345) and DDXPlus (+0.158). Detection is consistently the most discriminating metric: even where
fidelity scores appear competitive, detection reveals differences invisible to marginal and pairwise
statistics.

Privacy scores should be interpreted alongside the other metrics, as a model generating random data would 
achieve high privacy scores while failing on fidelity, utility and detection. For example, Tabby and CTGAN achieve perfect privacy while yielding poor detection scores. 
Except for one outlier, all methods yield privacy scores 
above 0.83 with DCR values near the ideal 50\%, indicating no systematic memorization. On Electric Vehicles, REalTabFormer's privacy score 
drops to 0.16, signaling a potential overfitting issue on the training data. \origami's privacy scores are all above 0.96. \origami produces a 
notable number of exact matches on Electric Vehicles (2,800 $\approx$ 1.5\%) and DDXPlus (1023 $\approx$ 0.1\%), and TabularARGN on DDXPlus 
(1,760 $\approx$ 0.2\%). However, we observe nearly identical exact match counts against the held-out test set (2,645 and 1,017, 
respectively for \origami), which the models never saw during training. This symmetry indicates that these matches arise from low-entropy records, 
common attribute combinations that recur throughout the dataset, rather than memorization of training examples.

\paragraph{Model sizes and training times}

Table~\ref{tab:training-cost} compares model sizes and training costs on the Yelp dataset for all models that could successfully train. \origami is the smallest model by a wide margin at 1.7M parameters, over 5$\times$ smaller than the second-smallest TabularARGN (9.0M) and 35$\times$ smaller than REalTabFormer (59.4M). Both RealTabFormer and TabDiff exceeded the 24-hour training budget. TabularARGN achieves the shortest total training time (6.3h vs.\ 13.3h for \origami), which can be attributed to its NADE-inspired MLP architecture that avoids the quadratic attention cost of transformers. 

\subsection{Preprocessing Artifacts}
\label{sec:preprocessing-artifacts}

\paragraph{Artifacts in numeric columns}

\begin{figure}[tb]
  \centering
  \includegraphics[width=\linewidth]{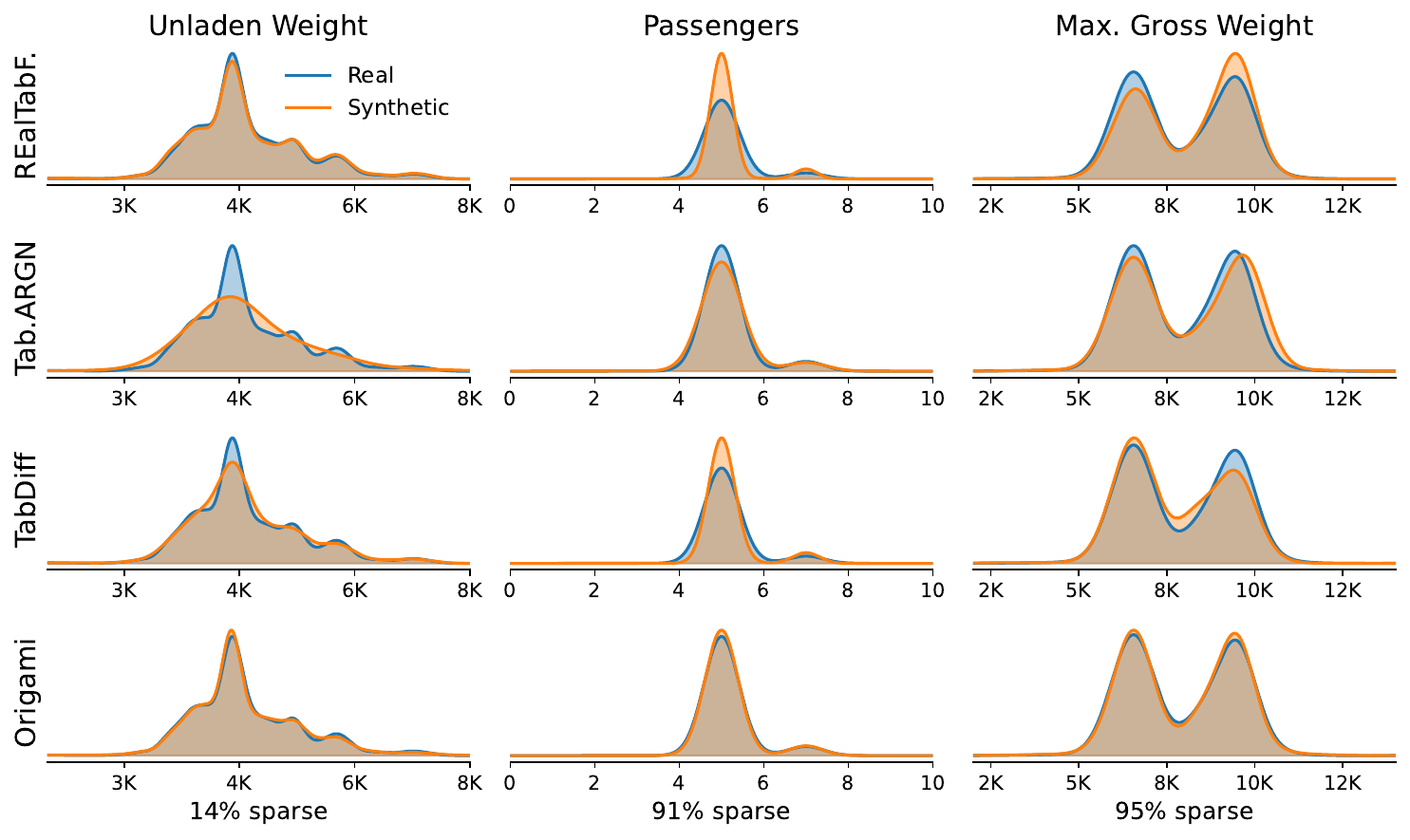}
  \caption{KDE visualizations of sparse numeric columns on the Electric Vehicles dataset. }
  \label{fig:kde-ev-plots}
\end{figure}

We further investigate notable outliers in the results by examining the Electric Vehicles dataset, where baseline detection scores start to degrade. Inspecting the feature importances of the XGBoost detection classifier reveals that the largest contributions come from numeric columns with high rates of structural sparsity: Maximum Gross Weight (95\% missing), Passengers (91\%), and Unladen Weight (14\%).
Figure~\ref{fig:kde-ev-plots} visualizes the kernel density estimates for these columns for all models that could train. For the two mostly sparse columns, Passengers and Max. Gross Weight, TabDiff shows clear over-estimation at the column mean. We attribute this to the mean imputation applied during TabDiff's preprocessing. When 95\% of a column's values are replaced by the column mean before training, the diffusion model learns a distribution dominated by this artificial mode rather than the true conditional distribution of the observed values. This highlights a fundamental limitation of mean imputation for structurally sparse data, where missingness is informative and the observed values follow a distribution far from the column mean.

For TabularARGN, the Unladen Weight column shows visible over-smoothing despite its low sparsity (14\%). TabularARGN encodes this column using quantile-based binning with uniform intra-bin sampling. Further analysis reveals that the sparse tail of the distribution is covered by a single bin spanning over 71,000 units, replacing a few clustered outliers with a uniform fill. In the dense core, the real data concentrates at a small number of exact vehicle weights, but the uniform sampling spreads these peaks across each bin's range, producing the smoother, flatter distribution visible in Figure~\ref{fig:kde-ev-plots}, again suggesting that numeric discretization can introduce artifacts that a good classifier can pick up on. 

\paragraph{Array Lengths.}

\begin{figure}[tb]
    \centering
    \includegraphics[width=\linewidth]{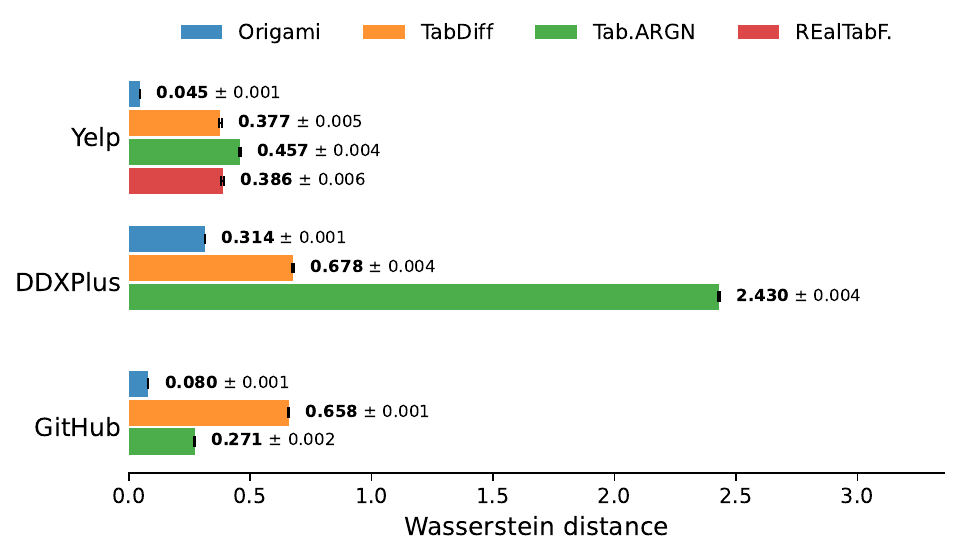}
    \caption{Wasserstein distance of length distributions between real and synthetic data on the Yelp ``\texttt{categories}'', DDXPlus ``\texttt{EVIDENCES}'' and Github Issues ``\texttt{issue.labels}'' arrays. Mean and variances of 3 seeds (lower is better).}
    \label{fig:wasserstein-yelp}
\end{figure}

A distinctive challenge for tabular baselines on semi-structured data is modeling array lengths. After flattening, each array element occupies a separate column, so length is only implicitly represented by the number of non-missing slots and is not directly optimized during training. To quantify this, we measure the Wasserstein distance \cite{villani2009optimal} between real and synthetic length distributions for Yelp \texttt{categories}, DDXPlus \texttt{EVIDENCES}, and GitHub Issues \texttt{issue.labels} arrays. \origami achieves substantially lower distances across all three arrays (Figure~\ref{fig:wasserstein-yelp}). This is consistent with our hypothesis: \origami generates tokens autoregressively while traversing the array and learns to terminate arrays naturally, whereas the tabular representation has no direct mechanism for capturing this joint constraint. This, however, is a limitation of the flattened representation rather than of the baseline models themselves: one could add explicit array-length features to the flattened tables, but doing so would provide side information that \origami does not receive, since it must learn array termination from the token sequence itself. Consequently, we chose not to provide it to the baseline models either. 

\subsection{Ablations}
\label{sec:ablations}

\paragraph{Position Encoding.}

To ablate the effectiveness of KVPE over sequential position encoding as used in GPT-2, we construct a pathological synthetic dataset of 5,000 nested JSON records with 4 branches (a–d), each containing two sub-objects (left, right) with a single boolean leaf key named val, for a total of 8 key paths per record that all share common key names: 

\begin{lstlisting}[language={}, basicstyle=\ttfamily\footnotesize]
{"a": {"left": {"val": true},  "right": {"val": false}},
 "b": {"left": {"val": false}, "right": {"val": true}},
 "c": {"left": {"val": true},  "right": {"val": false}},
 "d": {"left": {"val": true},  "right": {"val": false}}}
\end{lstlisting}

\begin{figure}[tb]
  \centering
  \begin{subfigure}[b]{0.47\linewidth}
    \centering
    \includegraphics[width=\textwidth]{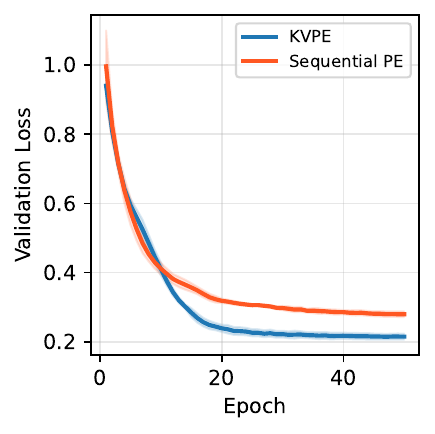}
    \caption{Validation loss}
    \label{fig:kvpe-val-loss}
  \end{subfigure}
  \hfill
  \hfill
  \begin{subfigure}[b]{0.5\linewidth}
    \centering
    \includegraphics[width=\textwidth]{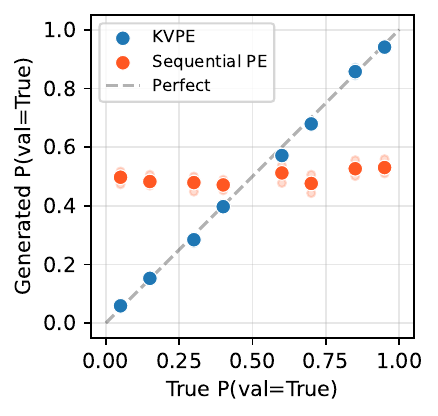}
    \caption{Sample statistics}
    \label{fig:kvpe-scatter}
  \end{subfigure}
  \caption{KVPE vs.\ sequential position encoding on a synthetic nested JSON dataset. KVPE accurately recovers path-specific marginals while sequential PE collapses to uniform ($\approx 0.5$). Results over 3 seeds.}
  \label{fig:kvpe-ablation}
\end{figure}

Each key path has a distinct Bernoulli parameter assigned, ranging from 0.05 to 0.95, from which we draw its boolean values across records. We train small \origami models (2 layers, d\_model=32, 4 heads) for 50 epochs with KVPE and sequential PE respectively, and measure per-path mean absolute error (MAE) of the generated marginals, repeating the training and sampling process across 3 random seeds.  Figure~\ref{fig:kvpe-ablation}(a) shows that KVPE achieves consistently lower validation loss. Figure~\ref{fig:kvpe-ablation}(b) confirms the cause: KVPE recovers the true per-path marginals (MAE $0.042 \pm 0.035$), while sequential PE collapses to a uniform ${\approx}0.5$ for all paths (MAE $0.271 \pm 0.003$), unable to disambiguate identically-named leaves at different key paths.

\begin{figure}[tbh]
    \centering
    \includegraphics[width=0.96\columnwidth]{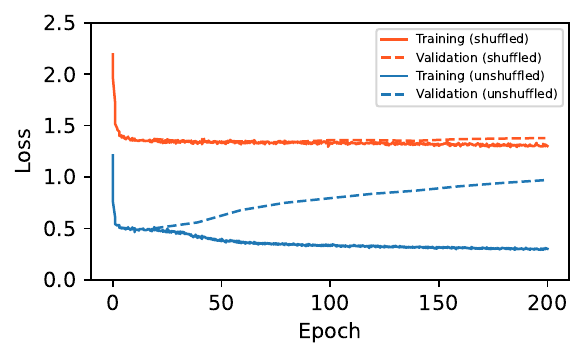}
    \caption{Training and validation loss curves on Adult with and without key shuffling. The unshuffled model starts to overfit at epoch 25 while the validation loss of the shuffled model remains stable. The higher loss baseline for the shuffled model is due to the unpredictability of randomized key tokens.}
    \label{fig:shuffle-ablation}
\end{figure}

\paragraph{Key Order Permutations.}

\origami and some other autoregressive methods \cite{borisovLanguageModelsAre2023, tiwaldTabularARGNFlexibleEfficient2025a} randomize the order of keys/columns in each record at every training step. We hypothesize that this prevents the model from exploiting key ordering as a spurious feature, forcing it to learn the actual statistical relationships between values.

We ablate this on the Adult dataset by training two identical models for 200 epochs: one with key shuffling enabled and one without.
The unshuffled model fits the training distribution slightly more tightly, yielding marginally higher fidelity (0.993 vs.\ 0.991), but this does not improve utility and comes at a steep privacy cost: DCR rises from 55.1\% to 72.8\%, and exact matches increase from 1 to 91. The loss curves in Figure~\ref{fig:shuffle-ablation} show the same pattern, with validation loss diverging after epoch 25 without shuffling (a classic overfitting signature). This supports key-order shuffling as an effective regularizer when combined with order-invariant position encoding.

%% file: 06_conclusion.tex
\section{Conclusion}

We presented \origami, an autoregressive transformer that synthesizes data directly
from tokenized JSON records, avoiding the flattening and imputation steps required by
tabular pipelines. By combining Key-Value Position Encoding, key-order shuffling, a
dual-head discrete/continuous architecture, and grammar and schema constraints, \origami
models nested objects, variable-length arrays, sparsity, and mixed-type values in the
record representation itself.

Across six datasets, \origami achieves the strongest fidelity and detection results and
is best or tied-best on utility in five of six cases, while maintaining privacy scores
above 96\%. On dense tabular benchmarks, the gains over strong baselines are moderate,
suggesting that these benchmarks are increasingly saturated. On JSON-native datasets,
the gains are larger, especially under the detection metric, where flattened
representations expose artifacts not captured by marginal and pairwise fidelity scores.

Overall, the results support a simple design lesson: for mixed-type sparse and semi-structured 
records, preserving structure and modeling missingness and arrays directly is a practical 
alternative to flatten-then-synthesize pipelines.

Several future directions remain open. Extending the architecture to multi-collection relational
data, where foreign-key dependencies introduce cross-record structure, is a natural next step.
More broadly, as an autoregressive model, \origami is a tractable density estimator over
semi-structured data, providing direct access to likelihood estimates. This opens
applications beyond synthetic data generation: conditional sampling and data imputation, 
cardinality estimation for query optimization---extending work on learned cardinality
estimators for relational tables~\cite{yangDeepUnsupervisedCardinality2019} to semi-structured 
data---predictive modeling over semi-structured records, and outlier detection via 
low-likelihood scoring.

%% file: references.bib
@misc{golkarXValContinuousNumber2023,
    title = {{xVal}: {A} {Continuous} {Number} {Encoding} for {Large} {Language} {Models}},
    url = {http://arxiv.org/abs/2310.02989},
    publisher = {arXiv},
    author = {Golkar, Siavash and Pettee, Mariel and Eickenberg, Michael and Bietti, Alberto and Cranmer, Miles and Krawezik, Geraud and Lanusse, Francois and McCabe, Michael and Ohana, Ruben and Parker, Liam and Blancard, Bruno R{\'e}galdo-Saint and Tesileanu, Tiberiu and Cho, Kyunghyun and Ho, Shirley},
    month = oct,
    year = {2023},
}

@article{gageNewAlgorithmData1994,
    title = {A new algorithm for data compression},
    volume = {12},
    number = {2},
    journal = {C Users J.},
    author = {Gage, Philip},
    month = feb,
    year = {1994},
    pages = {23--38},
}

@misc{wuGooglesNeuralMachine2016,
    title = {Google's {Neural} {Machine} {Translation} {System}: {Bridging} the {Gap} between {Human} and {Machine} {Translation}},
    url = {http://arxiv.org/abs/1609.08144},
    publisher = {arXiv},
    author = {Wu, Yonghui and Schuster, Mike and Chen, Zhifeng and Le, Quoc V. and Norouzi, Mohammad and Macherey, Wolfgang and Krikun, Maxim and Cao, Yuan and Gao, Qin and Macherey, Klaus and Klingner, Jeff and Shah, Apurva and Johnson, Melvin and Liu, Xiaobing and Kaiser, {\L}ukasz and Gouws, Stephan and Kato, Yoshikiyo and Kudo, Taku and Kazawa, Hideto and Stevens, Keith and Kurian, George and Patil, Nishant and Wang, Wei and Young, Cliff and Smith, Jason and Riesa, Jason and Rudnick, Alex and Vinyals, Oriol and Corrado, Greg and Hughes, Macduff and Dean, Jeffrey},
    month = oct,
    year = {2016},
}

@article{srivastavaDropoutSimpleWay2014,
    title = {Dropout: {A} {Simple} {Way} to {Prevent} {Neural} {Networks} from {Overfitting}},
    volume = {15},
    number = {56},
    urldate = {2024-10-22},
    journal = {Journal of Machine Learning Research},
    author = {Srivastava, Nitish and Hinton, Geoffrey and Krizhevsky, Alex and Sutskever, Ilya and Salakhutdinov, Ruslan},
    year = {2014},
    pages = {1929--1958},
}

@article{suRoFormerEnhancedTransformer2023a,
    title = {{RoFormer}: {Enhanced} Transformer with {Rotary} {Position} {Embedding}},
    volume = {568},
    doi = {10.1016/j.neucom.2023.127063},
    journal = {Neurocomputing},
    author = {Su, Jianlin and Ahmed, Murtadha and Lu, Yu and Pan, Shengfeng and Bo, Wen and Liu, Yunfeng},
    year = {2024},
    pages = {127063},
}

@inproceedings{vaswaniAttentionAllYou2017,
    title = {Attention is All You Need},
    booktitle = {Advances in Neural Information Processing Systems},
    volume = {30},
    author = {Vaswani, Ashish and Shazeer, Noam and Parmar, Niki and Uszkoreit, Jakob and Jones, Llion and Gomez, Aidan N. and Kaiser, {\L}ukasz and Polosukhin, Illia},
    year = {2017},
    pages = {5998--6008},
}

@misc{hendrycks2016gaussian,
    title = {Gaussian Error Linear Units ({GELUs})},
    url = {http://arxiv.org/abs/1606.08415},
    publisher = {arXiv},
    author = {Hendrycks, Dan and Gimpel, Kevin},
    year = {2016},
}

@inproceedings{pezoa2016foundations,
    title = {Foundations of {JSON} schema},
    booktitle = {Proceedings of the 25th international conference on world wide web ({WWW})},
    author = {Pezoa, Felipe and Reutter, Juan L. and Suarez, Fernando and Ugarte, Mart{\'\i}n and Vrgo{\v{c}}, Domagoj},
    year = {2016},
    pages = {263--273},
}

@misc{kellyUCIMLRepo,
    title = {The {UCI} {Machine} {Learning} {Repository}},
    url = {https://archive.ics.uci.edu},
    author = {Kelly, Markelle and Longjohn, Rachel and Nottingham, Kolby},
    institution = {University of California, Irvine, School of Information and Computer Sciences},
}

@inproceedings{shiTabDiffMixedtypeDiffusion2025,
    title={TabDiff: a Mixed-type Diffusion Model for Tabular Data Generation},
    author={Juntong Shi and Minkai Xu and Harper Hua and Hengrui Zhang and Stefano Ermon and Jure Leskovec},
    booktitle={The Thirteenth International Conference on Learning Representations},
    year={2025},
    url={https://openreview.net/forum?id=swvURjrt8z}
}

@article{pedregosaScikitlearnMachineLearning2011,
    title = {Scikit-learn: {Machine} learning in {Python}},
    volume = {12},
    journal = {Journal of Machine Learning Research},
    author = {Pedregosa, F. and Varoquaux, G. and Gramfort, A. and Michel, V. and Thirion, B. and Grisel, O. and Blondel, M. and Prettenhofer, P. and Weiss, R. and Dubourg, V. and Vanderplas, J. and Passos, A. and Cournapeau, D. and Brucher, M. and Perrot, M. and Duchesnay, E.},
    year = {2011},
    pages = {2825--2830},
}

@inproceedings{fansitchangoDdxplusNewDataset2022,
    title = {{DDXPlus}: {A} New Dataset for Automatic Medical Diagnosis},
    booktitle = {Advances in Neural Information Processing Systems},
    volume = {35},
    author = {Fansi Tchango, Arsene and Goel, Rishab and Wen, Zhi and Martel, Julien and Ghosn, Joumana},
    year = {2022},
    pages = {31306--31318},
}

@manual{sdmetrics,
    title = {Synthetic Data Metrics},
    organization = {DataCebo, Inc.},
    year = {2026},
    month = {02}, 
    note = {Version 0.12.0},
    url = {https://docs.sdv.dev/sdmetrics/}
}

@inproceedings{xuModelingTabularData2019,
    title = {Modeling Tabular Data using Conditional {GAN}},
    booktitle = {Advances in Neural Information Processing Systems},
    volume = {32},
    author = {Xu, Lei and Skoularidou, Maria and Cuesta-Infante, Alfredo and Veeramachaneni, Kalyan},
    year = {2019},
}

@inproceedings{kotelnikovTabDDPMModellingTabular2024,
    title = {{TabDDPM}: Modelling Tabular Data with Diffusion Models},
    booktitle = {International Conference on Machine Learning ({ICML})},
    author = {Kotelnikov, Akim and Baranchuk, Dmitry and Rubachev, Ivan and Babenko, Artem},
    year = {2023},
    pages = {17564--17579},
}

@inproceedings{zhangMixedTypeTabularData2024,
    title = {Mixed-Type Tabular Data Synthesis with Score-based Diffusion in Latent Space},
    booktitle = {International Conference on Learning Representations ({ICLR})},
    author = {Zhang, Hengrui and Zhang, Jiani and Srinivasan, Balasubramaniam and Shen, Zhengyuan and Qin, Xiao and Faloutsos, Christos and Rangwala, Huzefa and Karypis, George},
    year = {2024},
}

@misc{solatorioREaLTabFormerGeneratingRealistic2023,
    title = {{REaLTabFormer}: Generating Realistic Relational and Tabular Data using Transformers},
    url = {http://arxiv.org/abs/2302.02041},
    publisher = {arXiv},
    author = {Solatorio, Aivin V. and Dupriez, Olivier},
    month = feb,
    year = {2023},
}

@inproceedings{borisovLanguageModelsAre2023,
    title = {Language Models are Realistic Tabular Data Generators},
    booktitle = {International Conference on Learning Representations ({ICLR})},
    author = {Borisov, Vadim and Se{\ss}ler, Kathrin and Leemann, Tobias and Pawelczyk, Martin and Kasneci, Gjergji},
    year = {2023},
}

@article{crompTabbyLanguageModel2026,
    title = {Tabby: {A} Language Model Architecture for Tabular and Structured Data Synthesis},
    journal = {Transactions on Machine Learning Research},
    author = {Cromp, Sonia and GNVV, Satya Sai Srinath Namburi and Alkhudhayri, Mohammed and Cao, Catherine and Guo, Samuel and Roberts, Nicholas and Sala, Frederic},
    year = {2026},
    url = {https://openreview.net/forum?id=b9FPVnb0Bn},
}

@inproceedings{bourhisJSONDataModel2017a,
    title = {{JSON}: Data Model, Query Languages and Schema Specification},
    booktitle = {Proceedings of the 36th {ACM} {SIGMOD}-{SIGACT}-{SIGAI} Symposium on Principles of Database Systems ({PODS})},
    author = {Bourhis, Pierre and Reutter, Juan L. and Su{\'a}rez, Fernando and Vrgo{\v{c}}, Domagoj},
    year = {2017},
    pages = {123--135},
    doi = {10.1145/3034786.3056120},
}

@misc{shiComprehensiveSurveySynthetic2025,
    title = {A Comprehensive Survey of Synthetic Tabular Data Generation},
    url = {http://arxiv.org/abs/2504.16506},
    publisher = {arXiv},
    author = {Shi, Ruxue and Wang, Yili and Du, Mengnan and Shen, Xu and Chang, Yi and Wang, Xin},
    month = jul,
    year = {2025},
}

@misc{jordonSyntheticDataWhat2022,
    title = {Synthetic Data -- What, Why and How?},
    url = {http://arxiv.org/abs/2205.03257},
    publisher = {arXiv},
    author = {Jordon, James and Szpruch, Lukasz and Houssiau, Florimond and Bottarelli, Mirko and Cherubin, Giovanni and Maple, Carsten and Cohen, Samuel N. and Weller, Adrian},
    month = may,
    year = {2022},
}

@article{schmidtSQLStormTakingDatabase2025,
    title = {{SQLStorm}: Taking Database Benchmarking into the {LLM} Era},
    volume = {18},
    number = {11},
    journal = {Proceedings of the VLDB Endowment},
    author = {Schmidt, Tobias and Leis, Viktor and Boncz, Peter and Neumann, Thomas},
    year = {2025},
    pages = {4144--4157},
    doi = {10.14778/3749646.3749683},
}

@inproceedings{lopez-pazRevisitingClassifierTwoSample2018,
    title = {Revisiting Classifier Two-Sample Tests},
    booktitle = {International Conference on Learning Representations ({ICLR})},
    author = {Lopez-Paz, David and Oquab, Maxime},
    year = {2017},
}

@inproceedings{liuScalingPrivacyPreserving2024,
    title = {Scaling While Privacy Preserving: {A} Comprehensive Synthetic Tabular Data Generation and Evaluation in Learning Analytics},
    booktitle = {Proceedings of the 14th Learning Analytics and Knowledge Conference ({LAK})},
    author = {Liu, Qinyi and Khalil, Mohammad and Jovanovic, Jelena and Shakya, Ronas},
    year = {2024},
    pages = {620--631},
    doi = {10.1145/3636555.3636921},
}

@misc{tiwaldTabularARGNFlexibleEfficient2025a,
    title = {{TabularARGN}: {A} Flexible and Efficient Auto-Regressive Framework for Generating High-Fidelity Synthetic Data},
    url = {http://arxiv.org/abs/2501.12012},
    publisher = {arXiv},
    author = {Tiwald, Paul and Krchova, Ivona and Sidorenko, Andrey and Vieyra, Mariana Vargas and Scriminaci, Mario and Platzer, Michael},
    month = feb,
    year = {2025},
}

@inproceedings{uriaDeepTractableDensity2014,
    title = {A Deep and Tractable Density Estimator},
    booktitle = {International Conference on Machine Learning ({ICML})},
    author = {Uria, Benigno and Murray, Iain and Larochelle, Hugo},
    year = {2014},
    pages = {467--475},
}

@inproceedings{goodfellowGenerativeAdversarialNetworks2014,
    title = {Generative Adversarial Networks},
    booktitle = {Advances in Neural Information Processing Systems},
    volume = {27},
    author = {Goodfellow, Ian J. and Pouget-Abadie, Jean and Mirza, Mehdi and Xu, Bing and Warde-Farley, David and Ozair, Sherjil and Courville, Aaron and Bengio, Yoshua},
    year = {2014},
}

@inproceedings{kingmaAutoEncodingVariationalBayes2014,
    title = {Auto-Encoding Variational Bayes},
    booktitle = {International Conference on Learning Representations ({ICLR})},
    author = {Kingma, Diederik P. and Welling, Max},
    year = {2014},
}

@inproceedings{liu2023goggle,
    title = {{GOGGLE}: Generative Modelling for Tabular Data by Learning Relational Structure},
    booktitle = {International Conference on Learning Representations ({ICLR})},
    author = {Liu, Tennison and Qian, Zhaozhi and Berrevoets, Jeroen and van der Schaar, Mihaela},
    year = {2023},
}

@misc{radfordImprovingLanguageUnderstanding2018,
    title = {Improving Language Understanding by Generative Pre-Training},
    author = {Radford, Alec and Narasimhan, Karthik and Salimans, Tim and Sutskever, Ilya},
    year = {2018},
    url = {https://cdn.openai.com/research-covers/language-unsupervised/language_understanding_paper.pdf},
}

@inproceedings{yangXLNetGeneralizedAutoregressive2020,
    title = {{XLNet}: Generalized Autoregressive Pretraining for Language Understanding},
    booktitle = {Advances in Neural Information Processing Systems},
    volume = {32},
    author = {Yang, Zhilin and Dai, Zihang and Yang, Yiming and Carbonell, Jaime and Salakhutdinov, Ruslan and Le, Quoc V.},
    year = {2019},
}

@misc{alcornDEformerOrderAgnosticDistribution2021,
    title = {The {DEformer}: An Order-Agnostic Distribution Estimating Transformer},
    url = {http://arxiv.org/abs/2106.06989},
    publisher = {arXiv},
    author = {Alcorn, Michael A. and Nguyen, Anh},
    month = jul,
    year = {2021},
}

@inproceedings{patkiSyntheticDataVault2016,
    title = {The {Synthetic} {Data} {Vault}},
    booktitle = {2016 {IEEE} International Conference on Data Science and Advanced Analytics ({DSAA})},
    author = {Patki, Neha and Wedge, Roy and Veeramachaneni, Kalyan},
    year = {2016},
    pages = {399--410},
    doi = {10.1109/DSAA.2016.49},
}

@misc{willardEfficientGuidedGeneration2023,
    title = {Efficient {Guided} {Generation} for {Large} {Language} {Models}},
    url = {http://arxiv.org/abs/2307.09702},
    urldate = {2024-07-15},
    author = {Willard, Brandon T. and Louf, Rémi},
    month = aug,
    year = {2023},
    note = {arXiv:2307.09702 [cs]},
}

@inproceedings{kooAutomatabasedConstraintsLanguage2024,
    title = {Automata-Based Constraints for Language Model Decoding},
    url = {https://openreview.net/forum?id=BDBdblmyzY},
    booktitle = {Conference on Language Modeling},
    author = {Koo, Terry and Liu, Frederick and He, Luheng},
    year = {2024},
}

@article{yangDeepUnsupervisedCardinality2019,
  title     = {Deep Unsupervised Cardinality Estimation},
  author    = {Yang, Zongheng and Liang, Eric and Kamsetty, Amog and Wu, Chenggang and Duan, Yan and Chen, Xi and Abbeel, Pieter and Hellerstein, Joseph M. and Krishnan, Sanjay and Stoica, Ion},
  journal   = {Proceedings of the VLDB Endowment},
  volume    = {13},
  number    = {3},
  pages     = {279--292},
  year      = {2019},
  doi       = {10.14778/3368289.3368294}
}

@inproceedings{yangSAMDatabaseGeneration2022,
  title     = {{SAM}: Database Generation from Query Workloads with Supervised Autoregressive Models},
  author    = {Yang, Jingyi and Wu, Peizhi and Cong, Gao and Yang, Tong and Ruan, Jianfei},
  booktitle = {Proceedings of the 2022 International Conference on Management of Data},
  pages     = {1542--1555},
  year      = {2022},
  publisher = {ACM}
}

@article{gePrivBenchPrivacyenhancedDatabase2025,
  title   = {Privacy-Enhanced Database Synthesis for Benchmark Publishing},
  author  = {Ge, Yunqing and others},
  journal = {Proceedings of the VLDB Endowment},
  volume  = {18},
  number  = {2},
  year    = {2025}
}

@inproceedings{caiPrivLavaSynthesizingRelational2023,
  title     = {{PrivLava}: Synthesizing Relational Data with Foreign Keys under Differential Privacy},
  author    = {Cai, Kuntai and Xiao, Xiaokui and Cormode, Graham},
  booktitle = {Proceedings of the 2023 International Conference on Management of Data},
  pages     = {1--25},
  year      = {2023},
  publisher = {ACM}
}

@article{mohapatraDifferentiallyPrivateData2024,
  title   = {Differentially Private Data Generation with Missing Data},
  author  = {Mohapatra, Shubhankar and Zong, Jianqiao and Kerschbaum, Florian and He, Xi},
  journal = {Proceedings of the VLDB Endowment},
  volume  = {17},
  number  = {8},
  pages   = {2022--2035},
  year    = {2024}
}

@article{ostheimerSparseDataDiffusion2025,
  title   = {Sparse Data Diffusion for Scientific Simulations in Biology and Physics},
  author  = {Ostheimer, Phil and Nagda, Mayank and Balinskyy, Andriy and Radig, Jean and Herrmann, Carl and Mandt, Stephan and Kloft, Marius and Fellenz, Sophie},
  journal = {arXiv preprint arXiv:2502.02448},
  year    = {2025}
}

@misc{yuan_multi-faceted_2025,
    title = {A {Multi}-{Faceted} {Evaluation} {Framework} for {Assessing} {Synthetic} {Data} {Generated} by {Large} {Language} {Models}},
    url = {http://arxiv.org/abs/2404.14445},
    publisher = {arXiv},
    author = {Yuan, Yefeng and Liu, Yuhong and Cheng, Liang},
    month = jul,
    year = {2025},
}

@article{carey2025principled,
  author    = {Michael Carey and Wail Alkowaileet and Nick DiGeronimo and Peeyush Gupta and Sachin Smotra and Till Westmann},
  title     = {Towards Principled, Practical Document Database Design},
  journal   = {Proceedings of the VLDB Endowment},
  volume    = {18},
  number    = {12},
  pages     = {4804--4816},
  year      = {2025},
  doi       = {10.14778/3750601.3750606},
  url       = {https://www.vldb.org/pvldb/vol18/p4804-carey.pdf}
}

@article{stonebraker2024goes,
  title={What Goes Around Comes Around... And Around...},
  author={Stonebraker, Michael and Pavlo, Andrew},
  journal={ACM Sigmod Record},
  volume={53},
  number={2},
  pages={21--37},
  year={2024},
  publisher={ACM New York, NY, USA}
}

@book{villani2009optimal,
  title={Optimal transport: old and new},
  author={Villani, C{\'e}dric and others},
  volume={338},
  year={2009},
  publisher={Springer}
}
